\documentclass[11pt,a4paper]{article}
\usepackage{times,latexsym}
\usepackage{url}
\usepackage[T1]{fontenc}

\usepackage[acceptedWithA]{tacl2021v1}

\usepackage{xspace,mfirstuc,tabulary}

\newif\iftaclinstructions
\taclinstructionsfalse 
\iftaclinstructions
\renewcommand{\confidential}{}
\renewcommand{\anonsubtext}{(No author info supplied here, for consistency with
TACL-submission anonymization requirements)}
\newcommand{\instr}
\fi

\iftaclpubformat 

\else

\fi

\usepackage{hyperref}
\usepackage{url}
\usepackage{hyperref}       
\usepackage{url}            
\usepackage{booktabs}       
\usepackage{amsfonts}       
\usepackage{nicefrac}       
\usepackage{microtype}      
\usepackage{xcolor}         

\usepackage{url}
\usepackage{amsthm}
\usepackage{amsmath}
\usepackage{amssymb}
\usepackage{caption}
\usepackage{hyperref}
\usepackage{enumitem}
\usepackage{mathtools}
\usepackage{adjustbox}

\usepackage{amsmath}
\usepackage{amssymb}
\usepackage{mathtools}
\usepackage{amsthm}
\usepackage{caption}
\usepackage{bm}
\usepackage{multirow}
\usepackage[most]{tcolorbox}
\usepackage{xcolor}
\usepackage{enumitem}
\usepackage[most]{tcolorbox}
\usepackage{wrapfig}
\usepackage{xspace}
\newtcolorbox{takeawaybox}{
  colback=blue!3,
  colframe=blue!45!black,
  boxrule=0.6pt,
  arc=3.5pt,
  left=6pt,
  right=6pt,
  top=6pt,
  bottom=6pt,
  fonttitle=\bfseries
}
\newcommand{\ours}{\textsc{EFC-Adapter}\xspace}

\title{Scaling Laws for Agent Harnesses via Effective Feedback Compute}

\author{ Xuanliang Zhang \quad Dingzirui Wang \quad Keyan Xu \quad Qingfu Zhu \quad Wanxiang Che\Thanks{Corresponding author.}\\ Harbin Institute of Technology\\ \texttt{\{xuanliangzhang, dzrwang, kyxu, qfzhu, car\}@ir.hit.edu.cn} }

\date{}

\begin{document}
\maketitle
\begin{abstract}
Agent harnesses shape language-model performance by controlling tool use,
feedback, verification, memory, and repair.
Yet raw test-time expenditure, such as tokens, tool calls, wall time, or cost,
cannot distinguish useful feedback from redundant or unstable interaction.
We introduce \emph{Effective Feedback Compute} (EFC), a trace-level scaling
coordinate for informative, valid, non-redundant, and retained feedback.
We further define Estimated-EFC, NRS-EFC, harness efficiency $\eta$, and
task-demand normalization for realistic traces and heterogeneous tasks.
Across synthetic, real, held-out, and prospective evaluations,
EFC-based coordinates outperform raw-compute baselines and SAS.
Oracle-EFC/$D_{\mathrm{task}}$ reaches $R^2=0.99$ in controlled scaling, and
NRS-EFC/$D_{\mathrm{task}}$ reaches $R^2=0.93$ on real traces where raw compute
has near-zero or negative fit.
Finally, \ours uses EFC as a companion control layer for existing harnesses,
improving mean pass rate from $61.2\%$ to $68.2\%$ while reducing mean raw cost
from $213.8$ to $85.1$ under matched settings.
These results suggest that harness scaling depends on durable,
task-sufficient feedback rather than raw computation alone.
\end{abstract}


\section{Introduction}
    As large language models (LLMs) move from single-turn prediction to interactive problem
solving, performance increasingly depends on the \emph{agent harness}: the
machinery that routes tool calls, receives feedback, verifies intermediate
states, stores memory, repairs errors, and stops execution
\citep{yao2023react,schick2023toolformer,kim2026sas,ning2026codeagentharness}.
Harnesses therefore make test-time scaling a question of how inference-time
computation is converted into usable evidence from the environment
\citep{snell2025scaling,lee2026agentic,liu2025budget,kim2026agenticcoding}.
However, unlike pretraining, agent harnesses lack a clear scaling coordinate:
raw expenditure is insufficient because trajectories with the same tokens or
tool calls can differ sharply in whether their feedback is useful, valid,
non-redundant, and retained for later decisions
\citep{liu2025budget,lee2026agentic,kim2026agenticcoding,kim2026sas}.
This gap motivates our central question: \textit{what quantity should serve as
the scaling coordinate for closed-loop agent harness performance?}

We propose \emph{Effective Feedback Compute} (EFC) as this coordinate.
EFC credits feedback only when it is informative, valid, non-redundant, and
retained, thereby separating raw spending from feedback that can change future
behavior.
We further define harness efficiency
$\eta=\mathrm{EFC}/C_{\mathrm{raw}}$ to measure raw-to-EFC conversion, and
task-demand normalization $\mathrm{EFC}/D_{\mathrm{task}}$ to measure feedback
sufficiency relative to task demand.
For real execution traces with repeated or unstable observations, we use
non-redundant stable EFC (NRS-EFC) to emphasize retained feedback over transient
interaction.

We evaluate whether EFC explains harness scaling better than raw compute across
synthetic controllable tasks, semi-realistic executable code tasks, and real
benchmarks.
We compare EFC coordinates against raw tokens, tool calls, wall time,
operations, raw cost, and SAS \citep{kim2026sas}.
In controlled scaling, Oracle-EFC/$D_{\mathrm{task}}$ reaches $R^2=0.99$,
well above raw tokens, tool calls, and SAS.
Matched-budget interventions further show that improving feedback quality raises
success from $0.27$ to $0.90$ without changing raw cost or tool calls.
Trace-level experiments show that Estimated-EFC/$D_{\mathrm{task}}$
remains predictive without oracle-state access or final success labels.

We further show that EFC captures two complementary mechanisms in harness scaling.
Harness efficiency $\eta$ measures how routing, verification, memory, and
observation quality convert raw budget into effective feedback.
Task demand determines the scale on which this feedback becomes sufficient.
Across module ablations, cross-family prediction, mixed real traces, and a
prospective holdout, EFC-based coordinates consistently outperform raw-compute
baselines.
On real traces, NRS-EFC/$D_{\mathrm{task}}$ reaches $R^2=0.93$, and the
prospective holdout gives the strongest fit among the evaluated predictors
($R^2=0.89$).
Fitted task-demand calibration further improves transfer on mixed holdouts,
indicating that feedback sufficiency must be measured at the scale of the target
task distribution.
This decomposition makes the same coordinate useful for both predicting success
and diagnosing why a harness succeeds or fails on a given slice.
The efficiency analysis also shows that $\eta$ is slice-dependent, so harness
quality should be understood as a harness--task interaction rather than a fixed
property of the harness alone.


Finally, we test whether EFC can also guide harness control.
We introduce \ours, an EFC-aware companion layer for existing harnesses. 
\ours scores each available feedback action by its expected marginal gain in task-normalized EFC per unit raw cost, then uses this score to prioritize useful feedback, retain it in memory, gate self-evolution, and stop or roll back low-value loops. 
Across closed-loop agents, harness self-evolution, and trajectory-scaling methods, \ours improves the accuracy--cost tradeoff under the same model, task interface, action space, and raw-budget caps.

Our contributions are as follows: 
\begin{itemize}[nolistsep,leftmargin=*]
    \item We formalize EFC as a trace-level measure of useful feedback for closed-loop agent harnesses, with Estimated-EFC and NRS-EFC for settings without oracle state access. 
    \item We show that EFC and normalized EFC outperform raw-compute baselines and SAS across controlled, executable, real, and held-out evaluations. We further decompose harness scaling into raw-to-EFC conversion through harness efficiency and sufficiency relative to task demand. 
    \item We introduce \ours as an EFC-aware companion layer for existing harnesses. Under matched model, task, tool, action-space, and raw-budget-cap settings, \ours improves accuracy while reducing realized raw cost across closed-loop agents, harness self-evolution, and trajectory-scaling methods.
\end{itemize}

\section{Problem Formulation and Experimental Setup}
    \label{sec:preliminaries}
    We study whether the performance of agent harnesses can be explained by
a single scaling variable. Unlike standard inference-time scaling, where the main
resource is often a token budget or a number of samples, agent harnesses execute
closed-loop computations: they plan, act, observe external feedback, and update
their internal state. Our goal is to separate raw expenditure from feedback that
is actually useful for solving the task.

\subsection{Agent Harnesses as Closed-Loop Computation}
\label{sec:closed_loop}

Let $\mathcal{T}$ denote a task distribution. A task instance
$x \sim \mathcal{T}$ specifies an initial state, an instruction, an environment
interface, and an evaluation function. An agent harness $h \in \mathcal{H}$,
paired with a base model $m$, produces a trajectory
\begin{equation}
    \tau =
    \{(s_t, a_t, o_t, u_t)\}_{t=1}^{H},
\end{equation}
where $s_t$ is the agent state before step $t$, $a_t$ is a model action or tool call, $o_t$ is the resulting observation, and $u_t$ is the harness update to the agent state, memory, plan, or candidate solution. 
The horizon $H$ is determined by the harness stopping rule or by a budget limit.
Each run returns a final answer $\hat{y}$, which is evaluated by a task-specific checker $g_x(\hat{y}) \in \{0,1\}$. 
We define the expected success rate $S$ and error rate $E$ as
\begin{equation}
\begin{aligned}
    S(x,h,m,b) &= \mathbb{E}[g_x(\hat{y})], \\
    E(x,h,m,b) &= 1 - S(x,h,m,b).
\end{aligned}
\end{equation}
where $b$ denotes the raw budget configuration. 


\subsection{Evaluation Setup and Baselines}
\label{sec:evaluation_setup}

Full details about task construction, harness design, and budget accounting are provided in Appendix~\ref{app:detailed_experimental_setup}. 

\paragraph{Task layers.}
\label{sec:tasks}
We evaluate harness scaling on three task layers that progressively reduce
oracle access while preserving automatic evaluation. The \textbf{synthetic
controllable tasks} include Needle Lookup, State Tracking, and Rule Filter tasks
with hidden state and deterministic answers, which allow direct measurement of
Oracle-EFC and controlled variation in $D_{\mathrm{task}}$. The
\textbf{semi-realistic executable tasks} include code tasks and small executable
repair or analysis tasks with unit test or reference check feedback, which test
whether Estimated-EFC remains predictive when traces contain realistic model
errors. The \textbf{real benchmarks} include 
HumanEval~\citep{chen2021humaneval}, Terminal-Bench
2.0~\citep{merrill2026terminalbench}, and SWE-bench
Verified~\citep{jimenez2024swebench}, which test transfer to realistic agent trajectories.

\paragraph{Harness families.}
\label{sec:harnesses}
We compare seven harness families, denoted H0--H6, that differ in how they
convert raw budget into useful feedback. H0 is Direct Answer, H1 is
Checklist Verify, H2 is Routed Tools, H3 is
Stateful Memory, H4 is High Budget Noisy, H5 is
Closed Loop, and H6 is Deep Closed Loop. These harnesses vary
routing, verification, memory, interaction depth, and observation noise while
keeping the task distribution, final evaluator, model interface, and logging
protocol fixed within each experimental setting.

\paragraph{Models, budgets, and repeated runs.}
\label{sec:models_budgets}
We evaluate each task-harness configuration with \texttt{DeepSeek-V4-Flash}, \texttt{gpt-5.4-nano}, and \texttt{Claude-Haiku-4.5}. 
Unless stated otherwise, all reported numbers in the paper are first aggregated over repeated runs within each model and then averaged across models. 
We show the analysis results of each model in Appendix~\ref{app:model_specific_results}.

\paragraph{Scalar predictors and baselines.}
\label{sec:raw_baselines}
\label{sec:scalar_predictors}
For each run, we record raw tokens, tool calls, wall time, operations, and raw cost. 
We compare EFC predictors against these raw compute baselines and against SAS~\citep{kim2026sas}, a prior agent systems scaling baseline that uses a fixed effect equation over system level quantities. 

\section{Effective Feedback Compute}
    \label{sec:efc}
    We define Effective Feedback Compute (EFC) as a scalar measure of useful closed-loop feedback produced by an agent harness. 
A feedback event receives credit when it reveals task-relevant information, is grounded in reliable evidence, addresses the subgoal, and is retained for later decisions.

\subsection{Feedback Events}
\label{sec:feedback_events}

Given a trajectory $\tau = {(s_i, a_i, o_i, u_i)}_{i=1}^{H}$, where $H$ is the number of harness interaction steps, $s_i$ is the current state, $a_i$ is an action, $o_i$ is an observation, and $u_i$ is a state or memory update, we extract a sequence of atomic feedback events. 
A feedback event is a minimal unit of information in the trajectory that can confirm, reject, refine, or constrain the agent's subsequent decisions. 
After splitting compound observations and updates into such units, we denote the ordered feedback-event sequence extracted from $\tau$ by $\mathrm{Ev}(\tau)$: $\mathrm{Ev}(\tau) = (e_1,\ldots,e_T), T = |\mathrm{Ev}(\tau)|$.
Here $e_t$ is the $t$-th extracted feedback event, $t$ indexes feedback events, and $T$ is the number of extracted feedback events.

\subsection{Event-Level EFC}
\label{sec:event_level_efc}

Each event $e_t$ receives four bounded factors:
\begin{equation}
    I_t, V_t, R_t, M_t \in [0,1].
    \label{eq:efc_factors}
\end{equation}
Their meanings are as follows:
\begin{itemize}[nolistsep,leftmargin=*]
    \item \textbf{Informativeness $I_t$} reveals task-relevant information, such as a new constraint,
    reduced uncertainty, a diagnosed failure mode, or measurable subgoal progress.

    \item \textbf{Validity $V_t$} is supported by reliable evidence, such as a deterministic
    checker, execution result, unit test, or consistent tool observation.

    \item \textbf{Non-redundant relevance $R_t$} addresses the active subgoal and adds information beyond what is
    already available in the trajectory.

    \item \textbf{Memory update $M_t$}  changes the plan, state, or memory that can affect later actions.
\end{itemize}

The event contribution is
\begin{equation}
    \mathrm{EFC}_t
    =
    \kappa I_t V_t R_t M_t.
    \label{eq:event_efc}
\end{equation}
where $\kappa$ is a fixed scale constant. We keep $\kappa=10$ across all experiments.
The run-level EFC is
\begin{equation}
    \mathrm{EFC}(\tau)
    =
    \sum_{t=1}^{T}
    \mathrm{EFC}_t
    =
    \kappa
    \sum_{t=1}^{T}
    I_t V_t R_t M_t.
    \label{eq:efc}
\end{equation}
The product form gives high credit to feedback that is 
informative, valid, relevant, and retained.

\subsection{Oracle-EFC and Estimated-EFC}
\label{sec:oracle_estimated_efc}

For synthetic controllable tasks, hidden task state and ground-truth progress are available. 
We compute \emph{Oracle-EFC} by assigning $I_t,V_t,R_t,M_t$ from latent progress signals and deterministic checks.

For semi-realistic and real benchmark tasks, hidden task state is unavailable or
incomplete. We compute \emph{Estimated-EFC} from trace-observable features. Let
$\phi(e_t)$ denote the feature vector
\begin{equation}
    \phi(e_t)
    =
    [
    c_t,
    h_t,
    z_t,
    p_t,
    m_t,
    a_t,
    q_t,
    \Delta_t,
    \rho_t
    ],
    \label{eq:efc_trace_features}
\end{equation}
where $c_t$ indicates whether a checker fired, $h_t$ is checker scope, $z_t$
indicates whether a tool result is later referenced, $p_t$ indicates whether the
plan changes, $m_t$ measures memory retention, $a_t$ indicates repeated-error
avoidance, $q_t$ measures observation consistency, $\Delta_t$ measures subgoal
progress, and $\rho_t$ encodes trace position.

The event-level estimator is
\begin{equation}
    \widehat{\mathrm{EFC}}_t
    =
    \max
    \left(
    0,
    \exp
    \left(
    \theta_0 + \theta^{\top}\phi(e_t)
    \right)
    - 1
    \right),
    \label{eq:estimated_event_efc}
\end{equation}
and the run-level estimate is
\begin{equation}
    \widehat{\mathrm{EFC}}(\tau)
    =
    \sum_{t=1}^{T}
    \widehat{\mathrm{EFC}}_t.
    \label{eq:estimated_efc}
\end{equation}
The estimator is calibrated on controllable tasks with Oracle-EFC labels and then applied to traces without hidden-state access. 
The binary task outcome $g_x(\hat{y})$ is used only as the response variable when evaluating how well $\widehat{\mathrm{EFC}}(\tau)$ explains run-level success in the scaling fits.

For real execution traces, we also report status-aware variants. Let $Q_t$ be an observed status-quality score, $G_t$ a progress gate, $\Lambda_t$ a loop-type gate, and $A_t$ the attempt index. 
\textit{Estimated-EFC} is
\begin{equation}
    \widehat{\mathrm{EFC}}^{\mathrm{stable}}_t
    =
    \widehat{\mathrm{EFC}}_t Q_t G_t \Lambda_t.
    \label{eq:stable_estimated_efc}
\end{equation}
The nonredundant stable variant (\textit{NRS-EFC}) adjusts the event-level estimate to count feedback only insofar as it is reliable, nonredundant, stable, and not ambiguity-inducing:
\begin{equation}
\widehat{\mathrm{EFC}}^{\mathrm{nr}}_t
=
\frac{
\widehat{\mathrm{EFC}}_t
Q_t
G^{\mathrm{nr}}_t
\Lambda^{\mathrm{nr}}_t
}{
1 + \alpha_A A_t
}.
\label{eq:nonredundant_stable_efc}
\end{equation}
Here $\alpha_A$ controls the strength of the ambiguity penalty, with $\alpha_A=0.35$ across all experiments. 
The run-level NRS-EFC score for a trajectory is obtained by summing the event-level NRS-EFC:
\begin{equation}
\widehat{\mathrm{EFC}}^{\mathrm{nr}}(\tau)
=
\sum_{t=1}^{T}
\widehat{\mathrm{EFC}}^{\mathrm{nr}}_t .
\label{eq:run_level_nrs_efc}
\end{equation}


\subsection{Task Demand and Normalized EFC}
\label{sec:efc_normalization}


To compare tasks with different demands, we normalize EFC by a positive task-demand scale:
\begin{equation}
\begin{aligned}
    D_{\mathrm{task}}
    &=
    \max\left\{
    \epsilon_D,\right. \\
    &\left.
    L
    \cdot
    H_{\mathrm{tool}}
    \cdot
    S_{\mathrm{state}}
    \cdot
    (1 + N_{\mathrm{obs}})
    \cdot
    (1 - V_{\mathrm{ver}})
    \right\}.
\end{aligned}
\label{eq:dtask}
\end{equation}
Here $\epsilon_D>0$ prevents division by zero after normalization. $L$ is the
estimated minimum number of reasoning or action steps, $H_{\mathrm{tool}}$
measures tool-selection ambiguity, $S_{\mathrm{state}}$ measures state-tracking
demand, and $N_{\mathrm{obs}}$ measures observation noise or ambiguity.
$V_{\mathrm{ver}}$ denotes verifier-signal visibility: the extent to which the
task is covered by reliable checks, explicit tests, partial evaluators, or other
task-level validation signals. This term is an availability measure, which only
lowers $D_{\mathrm{task}}$ when reliable verification signals are more available.
All normalization constants and fitted demand exponents are estimated on the
calibration split and then frozen for held-out and prospective evaluations.

The normalized variables are
\begin{equation}
    X
    =
    \frac{\mathrm{EFC}}{D_{\mathrm{task}}},
    \qquad
    \widehat{X}
    =
    \frac{\widehat{\mathrm{EFC}}}{D_{\mathrm{task}}}.
    \label{eq:normalized_efc}
\end{equation}
We also report EFC efficiency:
\begin{equation}
    \eta
    =
    \frac{\mathrm{EFC}}{C_{\mathrm{raw}}},
    \qquad
    \widehat{\eta}
    =
    \frac{\widehat{\mathrm{EFC}}}{C_{\mathrm{raw}}},
    \label{eq:eta}
\end{equation}
where $C_{\mathrm{raw}}(\tau)$ is the realized raw-compute cost of trajectory $\tau$, defined from trace-observable resource usage and detailed in Appendix~\ref{app:raw_cost}. 

\subsection{Scaling Model and Evaluation Metrics}
\label{sec:scalar_metrics}

All scaling analyses use the same power-law failure model over a scalar
predictor $z$:
\begin{equation}
    E(z)
    =
    E_{\infty} + A z^{-\alpha},
    \label{eq:efc_scaling}
\end{equation}
where $E(z)$ is the predicted failure rate, $E_{\infty}$ is irreducible error, $A$ is a scale parameter, and $\alpha$ is the scaling exponent. 
We fit this model to raw-compute baselines and to EFC-based predictors. 
For scalar predictors, we median-normalize $z$ before fitting to put predictors with different units on a common numerical scale. 
Our primary EFC coordinates are raw EFC and their task-demand-normalized forms,
\begin{equation}
    X = \frac{\mathrm{EFC}}{D_{\mathrm{task}}},
    \qquad
    \widehat{X} = \frac{\widehat{\mathrm{EFC}}}{D_{\mathrm{task}}}.
    \label{eq:normalized_efc_metrics}
\end{equation}

We report two evaluation metrics: $R^2$ and MAE. 
Repeated runs are aggregated into the evaluation groups of each experiment.
On these aggregated groups, 
$R^2$ is computed as: $
R^2 = 1 - \left(\sum_i (\bar{E}_i-\widehat{E}_i)^2\right) / \left(\sum_i (\bar{E}_i-\bar{E})^2\right)
,$
where $\bar{E}_i$ is the observed failure rate of group $i$, $\widehat{E}_i$ is the predicted failure rate for that group, and $\bar{E}$ is the mean observed failure rate over the evaluated groups. 
Values closer to $1$ indicate better fit, and negative $R^2$ indicates that the predictor performs worse than predicting the mean failure rate. 
We also report mean absolute error, where lower values indicate smaller error: $\mathrm{MAE}
=
\frac{1}{n}
\sum_{i=1}^{n}
\left|
\bar{E}_i-\widehat{E}_i
\right|.
\label{eq}$

\begin{figure*}[t]
    \centering
    \includegraphics[width=.9\linewidth]{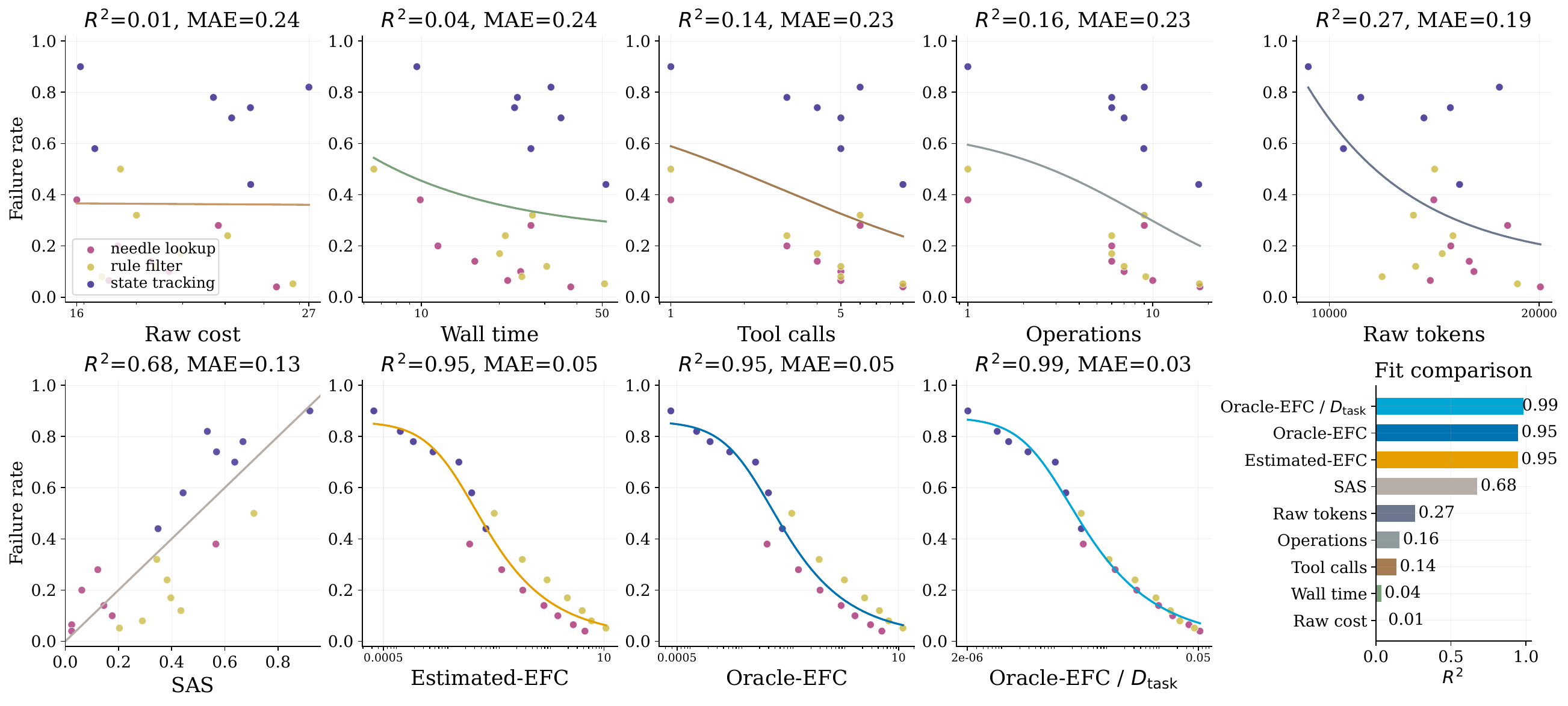}
    \caption{
    \textbf{Controlled scaling comparison on synthetic tasks.}
    The first nine panels, read left to right and top to bottom, fit the common power-law failure model to tool calls, operations, raw cost, wall time, raw tokens, SAS, Oracle-EFC, Estimated-EFC, and Oracle-EFC/$D_{\mathrm{task}}$.
    Points are aggregated failure rates, and curves are fitted trends. 
    The final panel summarizes the $R^2$ values, showing that EFC-based coordinates substantially outperform baselines and that Oracle-EFC/$D_{\mathrm{task}}$ gives the strongest curve collapse.
    }
    \label{fig:synthetic_scaling}
\end{figure*}

\section{Identifying EFC as a Scaling Coordinate}
    \label{sec:scaling_laws}
    This section tests whether EFC provides a scaling coordinate for agent harnesses. 
We first compare EFC-based coordinates against raw-compute scalars in controlled tasks, then use matched-budget comparisons to isolate feedback quality, and test whether EFC can be estimated from observable trajectory features without hidden-state access.

\subsection{Controlled Scaling Separates EFC from Raw Compute}
\label{sec:controlled_scaling}

We begin with synthetic controllable tasks, where Oracle-EFC can be measured
from hidden state and deterministic checks. For each task family, we evaluate
multiple harnesses under different budget levels and compare scalar coordinates
under the common power-law failure model in Eq.~\ref{eq:efc_scaling}. 
This experiment uses the full controlled set for descriptive scaling analysis. Its goal is to test whether Oracle-EFC/$D_{\mathrm{task}}$ separates effective feedback from raw expenditure in an idealized setting where the feedback coordinate is directly observable.

Figure~\ref{fig:synthetic_scaling} separates three increasingly informative
descriptions of a trajectory.
\emph{(i)} Raw-compute summaries are limited scaling coordinates. Raw cost, wall-clock time, tool calls, and operations explain only a small fraction of the
failure-rate variation, while raw tokens
are the strongest raw baseline but still reach only $R^2=0.27$. Thus, raw
expenditure and interaction counts capture some coarse budget differences but
cannot distinguish useful feedback from unproductive spending. \emph{(ii)}
Feedback-aware coordinates capture the missing signal. SAS improves sharply over
raw-compute summaries ($R^2=0.68$), while Oracle-EFC and Estimated-EFC both reach
$R^2=0.95$ with MAE $0.05$. 
The close match between Estimated-EFC and Oracle-EFC indicates that the useful-feedback signal can be recovered from observable trajectory features rather than from hidden state.
\emph{(iii)} Task normalization removes the remaining scale mismatch. 
Oracle-EFC/$D_{\mathrm{task}}$ reaches $R^2=0.99$ and MAE $0.03$, supporting the view that the relevant scaling coordinate is not absolute feedback alone, but feedback measured relative to task demand.

\subsection{Matched Budgets Probe Feedback Quality}
\label{sec:matched_budget}

A high-EFC trajectory might appear better simply because it spends more raw
compute. To reduce this confound, we construct matched-budget pairs on the same
task and model. The two conditions share the same token budget, tool-call
budget, wall-clock budget, operation count, and raw-cost accounting, while the
feedback returned to the harness differs in quality. The low-quality condition
produces noisy, redundant, and weakly retained observations, whereas the
high-quality condition produces targeted, valid, non-redundant feedback that
updates the agent state.

\begin{figure}[t]
    \centering
    \includegraphics[width=\linewidth]{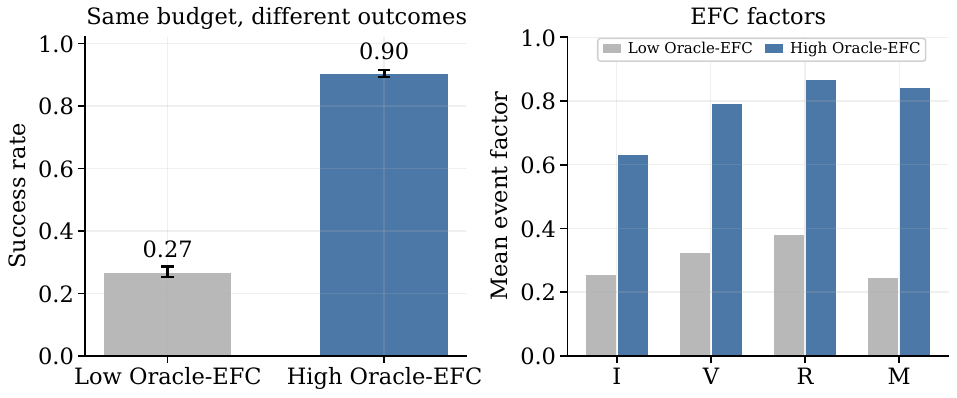}
    \caption{
    \textbf{Matched-budget feedback-quality control.}
    Left: under the same budget, the high-EFC condition increases success from $0.27$ to $0.90$. 
    Right: the high-EFC condition raises all EFC factors: informativeness (I), validity (V), non-redundancy (R), and memory update (M).
    }
    \label{fig:matched_budget}
\end{figure}

Figure~\ref{fig:matched_budget} provides a matched-budget comparison of feedback
quality. \emph{(i)} Success changes even when raw budget is held fixed. The low-
and high-feedback-quality trajectories are matched pairwise in raw budget, with
mean absolute raw-cost delta $0.000\%$ and mean tool-call delta $0.000$, yet the
high-feedback-quality condition increases success from $0.27$ to $0.90$
($p<10^{-6}$). This reduces the concern that the gain is driven by higher raw
compute, although it does not rule out all differences between trajectory types.
\emph{(ii)} The gain is associated with better raw-to-EFC conversion.
Informativeness, validity, non-redundancy, and memory update all increase
together, which matters because Eq.~\ref{eq:event_efc} has bottleneck behavior.
Feedback contributes little if it is invalid, redundant, irrelevant, or not
retained. Thus, the same raw budget succeeds when it is converted into
higher-quality effective feedback.

\subsection{Trace-Level Estimation Recovers Oracle-EFC}
\label{sec:trace_estimation}

Oracle-EFC requires hidden-state access and is therefore available only in
controllable environments. 
For non-oracle trajectories, EFC is estimated from logged trace features that exclude hidden state and the final success label.
We train an event-level Estimated-EFC model on synthetic calibration tasks and
evaluate it on held-out task families that are not used for estimator calibration.
The final success label is not used as an input.

\begin{figure}[t]
    \centering
    \includegraphics[width=.95\linewidth]{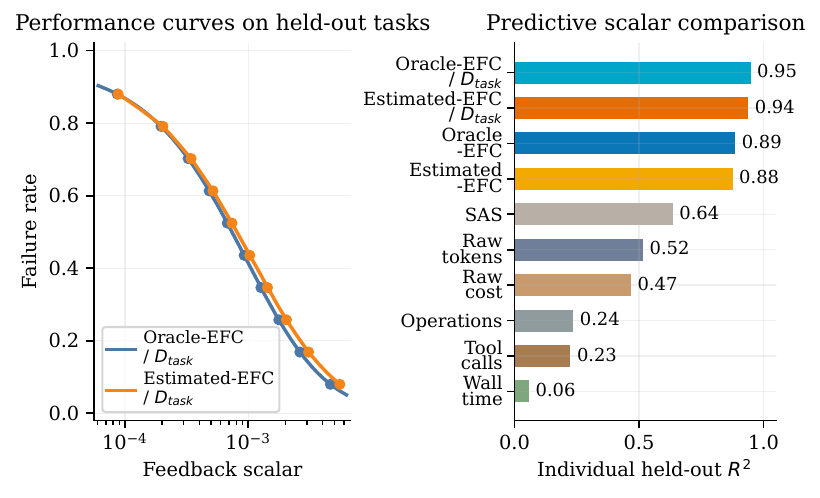}
    \caption{
    \textbf{Trace-level Estimated-EFC on held-out tasks.}
    Left: Estimated-EFC/$D_{\mathrm{task}}$ closely follows the held-out failure
    curve of Oracle-EFC/$D_{\mathrm{task}}$. 
    Right: the held-out $R^2$ ranking shows that the trace-level estimator nearly matches the oracle coordinate and outperforms raw compute and SAS.
    }
    \label{fig:estimated_efc}
\end{figure}


Figure~\ref{fig:estimated_efc} addresses whether EFC is only an oracle
diagnostic. \emph{(i)} Estimated-EFC recovers most of the oracle feedback
coordinate without hidden-state access. Without the task normalization $D_{\mathrm{task}}$,
Estimated-EFC reaches $R^2=0.88$, close to Oracle-EFC at $0.89$.
\emph{(ii)} Demand normalization improves oracle and estimated coordinates in
the same direction. Estimated-EFC/$D_{\mathrm{task}}$ and
Oracle-EFC/$D_{\mathrm{task}}$ rise to $0.94$ and $0.95$, while raw-compute
baselines remain much weaker, ranging from wall time at $0.06$ to raw tokens at
$0.52$. This matched improvement shows that trace-observable evidence captures
useful feedback, and that $D_{\mathrm{task}}$ removes residual task-family scale
differences rather than merely re-ranking runs after the fact.


\section{Decomposing EFC: Harness Efficiency and Task Demand}
    \label{sec:mechanisms_difficulty}
Section \ref{sec:scaling_laws} shows that EFC is a stronger scaling coordinat
e than raw compute.
We next decompose EFC into two mechanisms: harness efficiency $\eta$, which measures effective feedback per unit raw budget, and task-normalized feedback $\mathrm{EFC}/D_{\mathrm{task}}$, which measures sufficiency relative to task demand.
Thus, harness design governs raw-to-EFC conversion through $\eta$, while $D_{\mathrm{task}}$ sets the scale on which feedback becomes sufficient.

\subsection{Harness Factors Control Raw-to-EFC Conversion}
\label{sec:factor_efficiency}

We first vary individual harness and environment factors in controlled synthetic
tasks while keeping the task distribution and budget accounting fixed.
For each factor, we compare low, medium, and high levels and measure the induced change in harness efficiency $\eta$ and the corresponding success rate.

\begin{figure}[t]
    \centering
    \includegraphics[width=.95\linewidth]{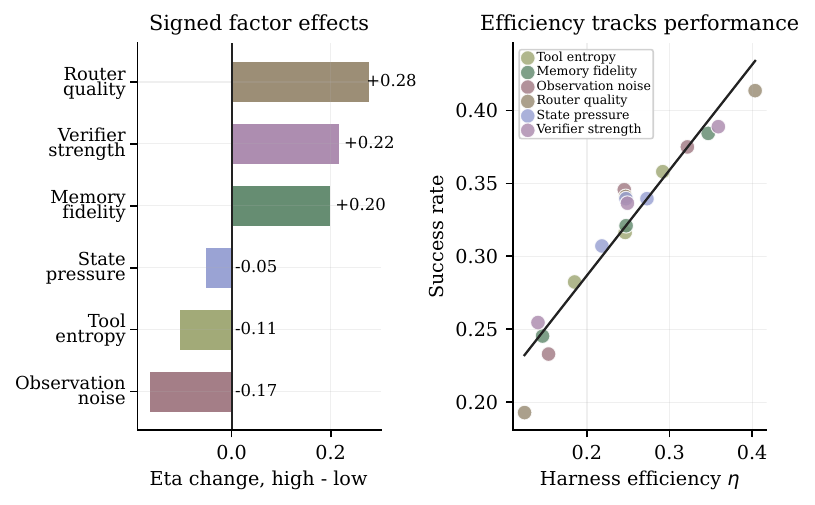}
    \caption{
    \textbf{Harness factors control raw-to-EFC conversion.}
    Left: signed high-minus-low effects on $\eta$ show that router quality, verifier strength, and memory fidelity increase $\eta$, and observation noise, tool entropy, and state pressure reduce it. 
    Right: success rate tracks $\eta$ across factor settings, indicating that these design and interface changes affect performance primarily by changing how much effective feedback is extracted from the same budget.
    }
    \label{fig:eta_decomposition}
\end{figure}

Figure~\ref{fig:eta_decomposition} shows that the same raw budget can be converted into very different amounts of effective feedback. 
\emph{(i)} The largest gains come from factors that improve the selection and reliability of feedback. 
Router quality gives the largest efficiency gain ($+0.28$), followed by verifier strength and memory fidelity. 
This ordering is consistent with Eq.~\ref{eq:event_efc}: routing improves relevance and non-redundancy, verification improves validity, and memory improves retention.
\emph{(ii)} Interface frictions reduce efficiency by corrupting or diluting feedback. Observation noise produces the largest drop ($-0.17$), tool entropy also reduces efficiency, and state pressure has a smaller negative effect. 
\emph{(iii)} Outcomes track conversion rather than budget alone. 
Success rises approximately monotonically with $\eta$, from below $0.20$ to above $0.40$ at the most efficient settings.
Thus, these factors affect performance by changing the raw-to-EFC conversion rate.
Module ablations further confirm that $\eta$ mediates success, with details in Appendix~\ref{sec:module_ablation}.

\subsection{Task Demand Sets the Required EFC Scale}
\label{sec:difficulty_collapse}

A harness can convert raw budget into EFC efficiently and still fail when the
task requires more feedback than the trajectory provides. 
We therefore test whether task demand provides the scale needed to compare EFC across task families. 
In controlled tasks, we evaluate raw compute baselines, SAS, raw Oracle-EFC, and two demand-normalized Oracle-EFC coordinates under the same power-law failure model. 
The hand-designed demand follows Eq.~\ref{eq:dtask}.
The fitted variant uses the same task-demand factors but estimates their
relative exponents from calibration tasks.

\begin{figure}[t]
    \centering
    \includegraphics[width=.8\linewidth]{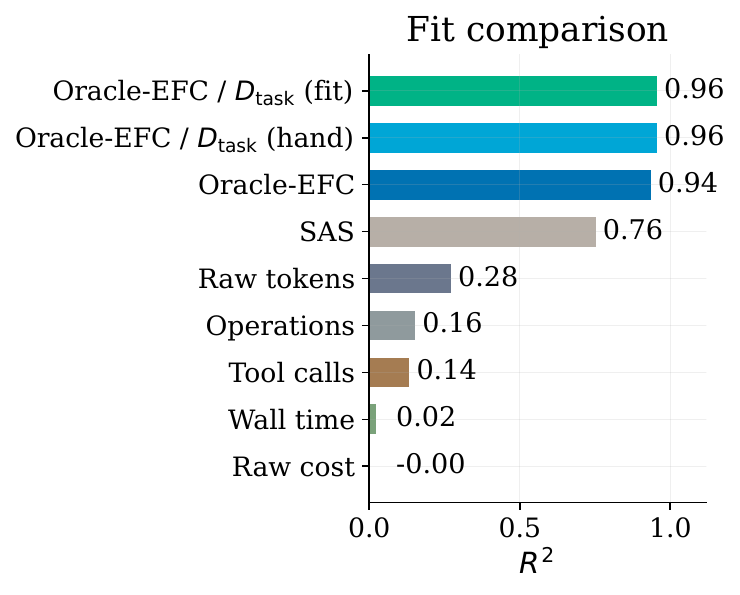}
    \caption{
    \textbf{Task demand sets the required EFC scale.}
    Bars summarize the $R^2$ values of raw compute, SAS, Oracle-EFC, Oracle-EFC normalized by hand-designed task demand, and Oracle-EFC normalized by fitted task demand.
    Raw-compute summaries capture only limited cross-family structure, while feedback-aware coordinates are substantially stronger and demand-normalized Oracle-EFC gives the best curve collapse.
    }
    \label{fig:difficulty_collapse}
\end{figure}

Figure~\ref{fig:difficulty_collapse} shows that task demand provides the missing scale factor for cross-family prediction.
\emph{(i)} Raw budget gives limited cross-family alignment.
Raw cost, wall time, tool calls, and operations reach only $R^2<0.20$.
Raw tokens improve to $R^2=0.28$, but still leave family-dependent offsets.
\emph{(ii)} Feedback-aware coordinates align the controlled task families better.
SAS reaches $R^2=0.76$, and Oracle-EFC reaches $0.94$, showing that trace structure and effective feedback explain more variation than raw resource use.
However, absolute feedback does not account for the feedback scale required by each task family.
\emph{(iii)} Normalizing EFC by task demand removes most residual scale mismatch.
Oracle-EFC/$D_{\mathrm{task}}$ reaches $R^2=0.96$ with the hand-designed denominator and $0.96$ with the fitted denominator.
The small fitting gain and the strong performance of both variants support that EFC is most predictive when measured relative to the task-specific feedback scale.


\subsection{Task-Demand Calibration Transfers to Mixed Holdout}
\label{sec:difficulty_transfer}

We test whether task-demand normalization transfers beyond the controlled task families.
We evaluate a mixed held-out set with heterogeneous tasks and compare baselines, NRS-EFC, hand-designed NRS-EFC/$D_{\mathrm{task}}$, and fitted NRS-EFC/$D_{\mathrm{task}}$. 
The fitted model uses the same factors as before, but learns their exponents on a calibration split and evaluates the coordinate on unseen tasks.

\begin{figure}[t]
    \centering
    \includegraphics[width=.95\linewidth]{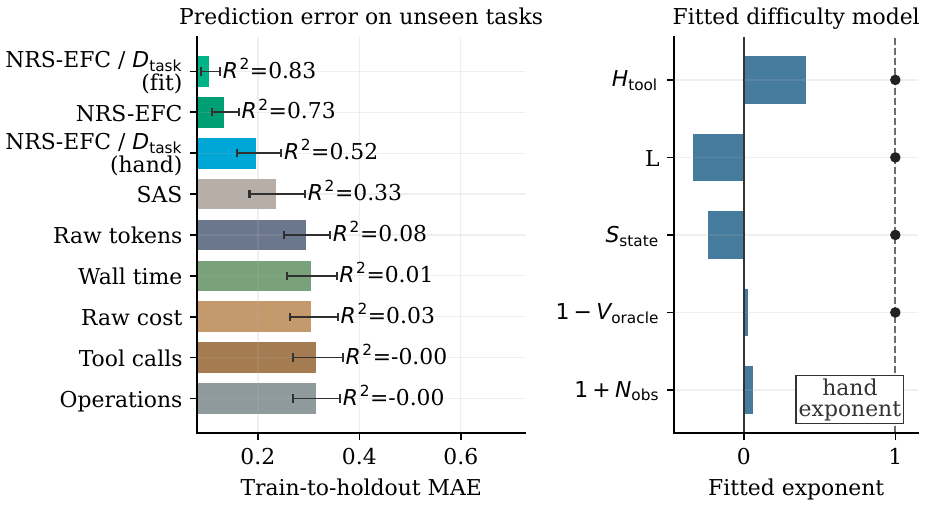}
    \caption{
    \textbf{Task-demand calibration transfers to mixed held-out tasks.}
    Left: prediction error on unseen tasks compares raw compute, SAS, NRS-EFC, hand-designed NRS-EFC/$D_{\mathrm{task}}$, and fitted
    NRS-EFC/$D_{\mathrm{task}}$. 
    Horizontal error bars denote task-level bootstrap $95\%$ confidence intervals for MAE.
    Right: the fitted task-demand model learns exponents over demand factors.
    }
    \label{fig:difficulty_transfer}
\end{figure}

Figure~\ref{fig:difficulty_transfer} shows that task-demand calibration improves the portability of the EFC coordinate.
\emph{(i)} Raw compute transfers weakly to unseen mixed tasks.
Raw tokens are the strongest raw baseline, with $R^2=0.08$, while all expenditure summaries are weak.
SAS is more stable, but remains limited.
\emph{(ii)} NRS-EFC transfers better than raw expenditure, and calibrated task demand transfers best.
The manually designed NRS-EFC/$D_{\mathrm{task}}$ is weaker on the heterogeneous holdout, whereas the fitted normalization achieves the best result with $R^2=0.83$.
\emph{(iii)} The fitted exponents indicate uneven demand contributions.
$H_{\mathrm{tool}}$ receives the largest weight, suggesting that tool selection ambiguity is the main residual difficulty after NRS-EFC is accounted for.
Other factors add smaller or negative corrections, so the fitted denominator should be read as a distribution-specific calibration of $D_{\mathrm{task}}$, not as a universal causal weighting.

\section{Held-Out and Prospective Validation}
    \label{sec:generalization}
The previous sections identify EFC as the scaling coordinate and decompose it into harness efficiency and task demand.
We next test whether this coordinate transfers beyond the fitted configurations.

\subsection{Task-Demand-Normalized EFC Predicts Unseen Configurations}
\label{sec:heldout_prediction}

We evaluate held-out prediction by removing configurations along four axes: task
family, harness variant, model, and combined setting. For each split, we fit the
power-law scaling model on the remaining configurations in failure-rate space.
We report the equivalent success-rate prediction $1-\widehat{E}(z)$ on the held-out group. 
The task-demand denominator is either hand-designed from task-demand factors or fitted on the training split. 

\begin{figure*}[t]
    \centering
    \includegraphics[width=.95\linewidth]{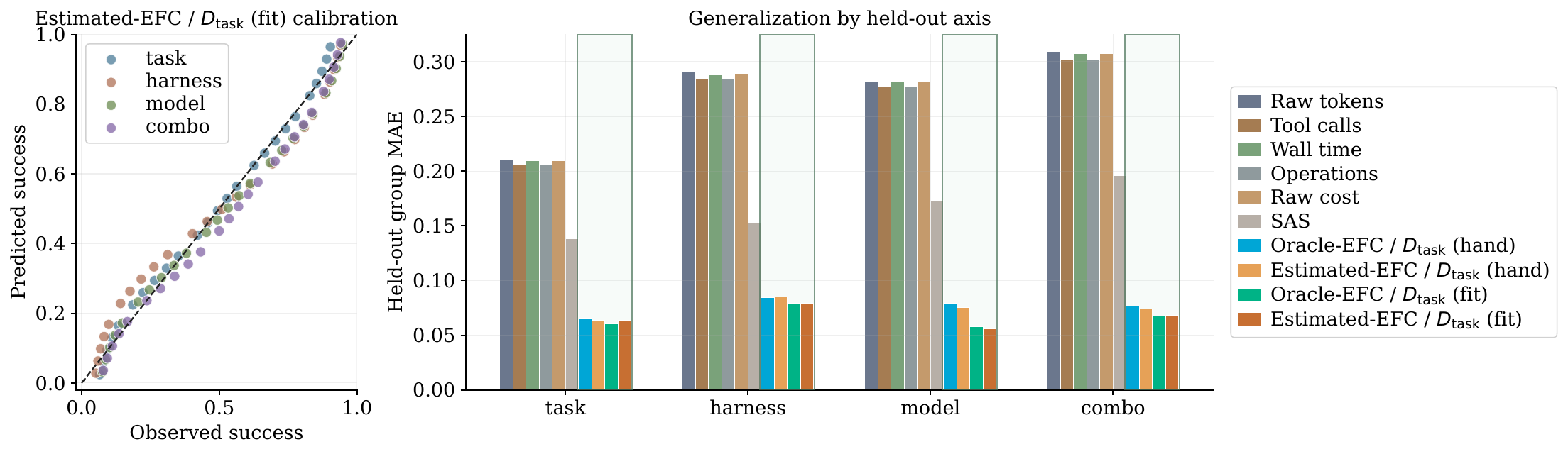}
    \caption{
    \textbf{Task-demand-normalized EFC predicts unseen configurations.}
    Left: predicted success, computed as $1-\widehat{E}(z)$ from the fitted
    failure-rate scaling model, is calibrated against observed success across
    held-out task, harness, model, and combined splits. Right: grouped bars report
    held-out group MAE for raw compute baselines, SAS, and task-demand-normalized
    EFC coordinates across the four held-out axes. Task-demand-normalized EFC gives
    the lowest prediction error across all axes.
    }
    \label{fig:heldout_prediction}
\end{figure*}

Figure~\ref{fig:heldout_prediction} shows that the EFC coordinate transfers
across unseen axes. \emph{(i)}
Calibration is preserved at the level of absolute success after converting the fitted failure-rate predictions to success rates.
The left panel stays
close to the diagonal across task, harness, model, and combined splits, which
indicates that fitted Estimated-EFC/$D_{\mathrm{task}}$ predicts success rates
rather than only ranking configurations. \emph{(ii)} The coordinate ordering is
stable across held-out axes. Raw-compute scalars have the largest errors, and task-demand-normalized EFC forms the lowest-error group on
every split. The gain is largest on harness and combined splits, where raw
expenditure is most confounded by changed decision policies. \emph{(iii)}
Fitted and hand-designed task demand are close. Their similar errors show that
the transfer signal mainly comes from measuring feedback relative to task demand,
while fitting improves calibration under distribution shift.

\subsection{Non-Redundant Stable EFC Reveals Slice-Specific Harness Efficiency}
\label{sec:real_mix}

We next evaluate whether the same feedback-based scaling relationship survives on
real execution traces, where observations are noisier, errors are often
repeated, and intermediate states are less controlled than in the synthetic settings. 
We therefore use NRS-EFC, which keeps feedback events that are both informative and retained, while down-weighting redundant or unstable signals that do not contribute durable progress. 
We pool heterogeneous real slices, and compare baselines, NRS-EFC, and NRS-EFC/$D_{\mathrm{task}}$ under the same aggregated scaling protocol.
We also examine $\eta$ using NRS-EFC in order to understand how different harnesses convert raw budget into effective feedback.

\begin{figure}[t]
    \centering
    \includegraphics[width=.95\linewidth]{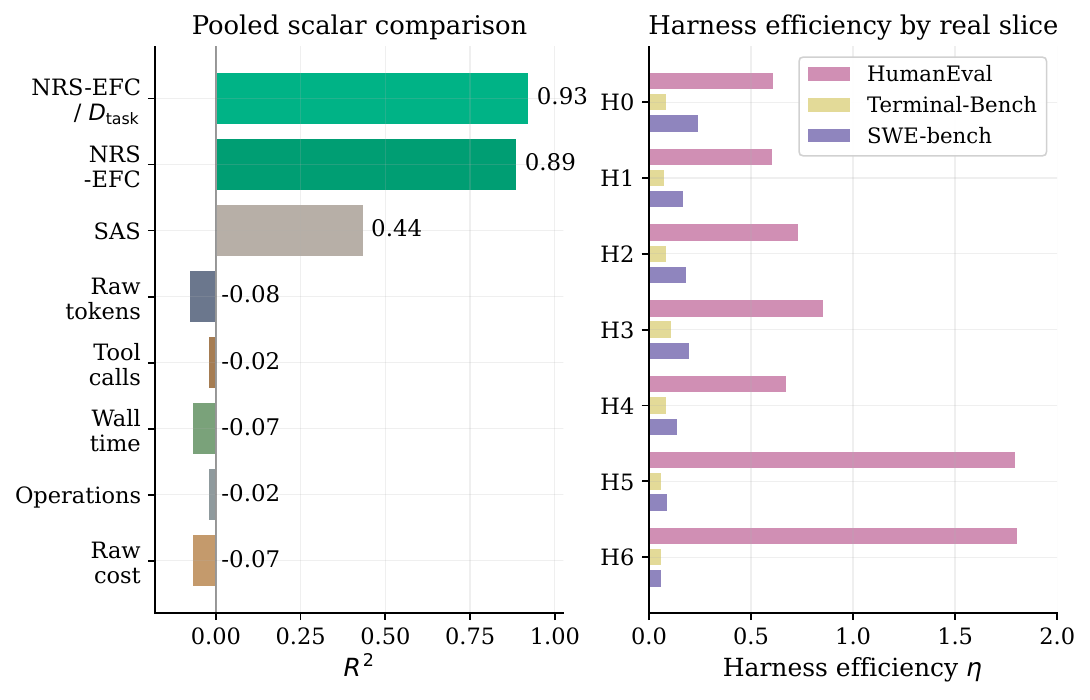}
    \caption{
    \textbf{NRS-EFC reveals slice-specific harness efficiency.}
    Left: pooled scalar comparison on mixed real traces, reporting predictive $R^2$.
    Demand-normalized NRS-EFC gives the strongest pooled fit.
    Right: $\eta$ is shown for H0--H6 across HumanEval, Terminal-Bench, and SWE-bench slices. 
    }
    \label{fig:real_mix}
\end{figure}

Figure~\ref{fig:real_mix} shows two main results.
\emph{(i)} Raw compute largely fails as a pooled predictor on heterogeneous real traces.
All raw expenditure summaries have negative $R^2$.
In contrast, NRS-EFC reaches $R^2=0.89$, and NRS-EFC/$D_{\mathrm{task}}$ further improves to $0.93$.
SAS is stronger than raw compute at $0.44$, but remains well below the feedback-based coordinates.
Thus, success is better predicted by nonredundant, retained feedback relative to task demand than by raw budget alone.
\emph{(ii)} Harness efficiency is strongly slice-specific.
On HumanEval, H5 and H6 achieve the highest $\eta$, suggesting that richer verification and feedback use are effective for executable coding tasks.
On Terminal-Bench, all harnesses remain near low $\eta$, indicating that feedback is difficult to convert into reusable progress.
On SWE-bench, earlier or mid-stage harnesses such as H0 and H3 are strongest, while H5 and H6 no longer dominate.
These reversals suggest that $\eta$ reflects a harness--task interaction rather than an invariant property of the harness alone.

\subsection{Prospective Holdout Validates the Scaling Coordinate} 
\label{sec:prospective_holdout} 

\begin{figure}[t]
\centering 
\vspace{-0.8em} 
\includegraphics[width=.85\linewidth]{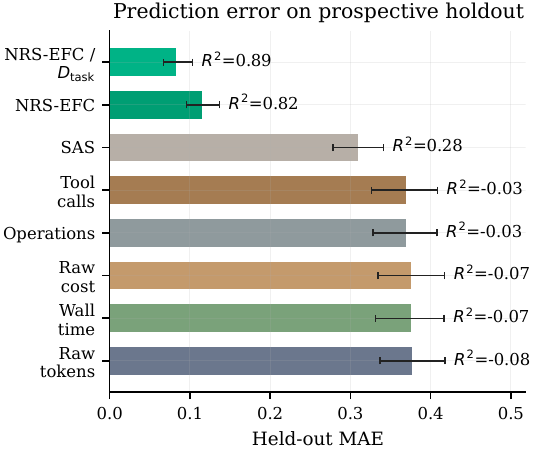} 
\caption{ 
\textbf{Prospective holdout prediction.} 
Bars report held-out MAE and $R^2$ on a prospective batch of unseen real traces, and horizontal error bars denote task-level bootstrap $95\%$ confidence intervals.
} 
\label{fig:prospective_holdout} 
\vspace{-0.8em} 
\end{figure} 

We conclude with a prospective holdout test that evaluates whether the same coordinate transfers to traces that were not available during metric design or calibration. Before collecting the prospective runs, we fixed the prediction protocol, including the definition of NRS-EFC, the task-demand factors, the fitted task-demand exponents, and all comparison baselines. We then evaluate a new held-out batch of real traces and compare raw compute baselines, SAS, NRS-EFC, and NRS-EFC/$D_{\mathrm{task}}$ using the same held-out prediction procedure as above. This setting asks whether a prespecified feedback-based coordinate remains predictive when both the task mix and the evaluation examples are unseen. 

Figure~\ref{fig:prospective_holdout} shows that the ordering observed in earlier validation settings transfers to the prospective batch. \emph{(i)} Feedback-based coordinates remain predictive on new real traces. NRS-EFC/$D_{\mathrm{task}}$ achieves the best held-out prediction, with the highest held-out $R^2$ of $0.89$, while raw NRS-EFC follows closely at $0.82$. SAS remains informative but is substantially weaker at $R^2=0.28$. \emph{(ii)} Raw expenditure is not a reliable out-of-sample coordinate. Raw tokens, tool calls, wall time, operations, and raw cost all obtain negative held-out $R^2$, indicating that they perform worse than a mean predictor on the prospective batch. \emph{(iii)} Task-demand normalization provides a targeted calibration gain rather than an evaluation-set-specific adjustment. NRS-EFC already captures nonredundant retained feedback, and dividing by $D_{\mathrm{task}}$ improves prediction when the prospective batch mixes tasks with different feedback requirements. Because the coordinate and calibration procedure were specified before evaluating the prospective batch, the improvement supports the proposed scaling coordinate rather than post hoc adaptation to the held-out examples.

    
\section{EFC-Adapter: An EFC-aware Companion Layer}
    \label{sec:methodology}

We now use the feedback coordinate as a control signal. Given a trajectory prefix
$\tau_{\le t}=\{(s_i,a_i,o_i,u_i)\}_{i=1}^{t}$, \ours attaches to a base harness
and estimates a causal prefix score
$\widehat{\mathrm{NRS\text{-}EFC}}(\tau_{\le t})/D_{\mathrm{task}}$.
The adapter preserves the same model, task interface, tool set, action space,
and raw-budget caps as the base harness.
At step $t$, \ours ranks available feedback actions by their expected marginal
increase in $\widehat{\mathrm{NRS\text{-}EFC}}(\tau_{\le t})$ per unit
$C_{\mathrm{raw}}$, retains high-scoring non-redundant feedback in memory, gates self-evolution updates, and stops or rolls back when additional raw cost yields
little estimated feedback gain. 
The prefix score uses only information in $\tau_{\le t}$, which omits
the future-reference feature $z_t$, replaces the retention feature $m_t$ with a
prefix-observable memory-write signal in Eq.~\ref{eq:efc_trace_features}.
Details of \ours is in Appendix~\ref{app:efc_adapter}.

We evaluate \ours as a matched-budget companion layer attached to four representative baselines, instantiating three lines of agent-harness work. 
\textbf{mini-SWE-agent}~\cite{yang2024sweagent} represents lightweight closed-loop coding agents that solve tasks through execution, observation, and iterative repair. 
\textbf{AHE}~\cite{lin2026ahe} represents harness self-evolution, using an evaluate--analyze--improve loop to update harness policies from observed failures. 
\textbf{RTV} and \textbf{PDR} represent test-time trajectory scaling: RTV recursively selects among sampled rollouts through comparison voting, and PDR conditions fresh attempts on summaries of prior rollouts \citep{kim2026agenticcoding}. 

\begin{figure}[t]
    \centering
    \includegraphics[width=\linewidth]{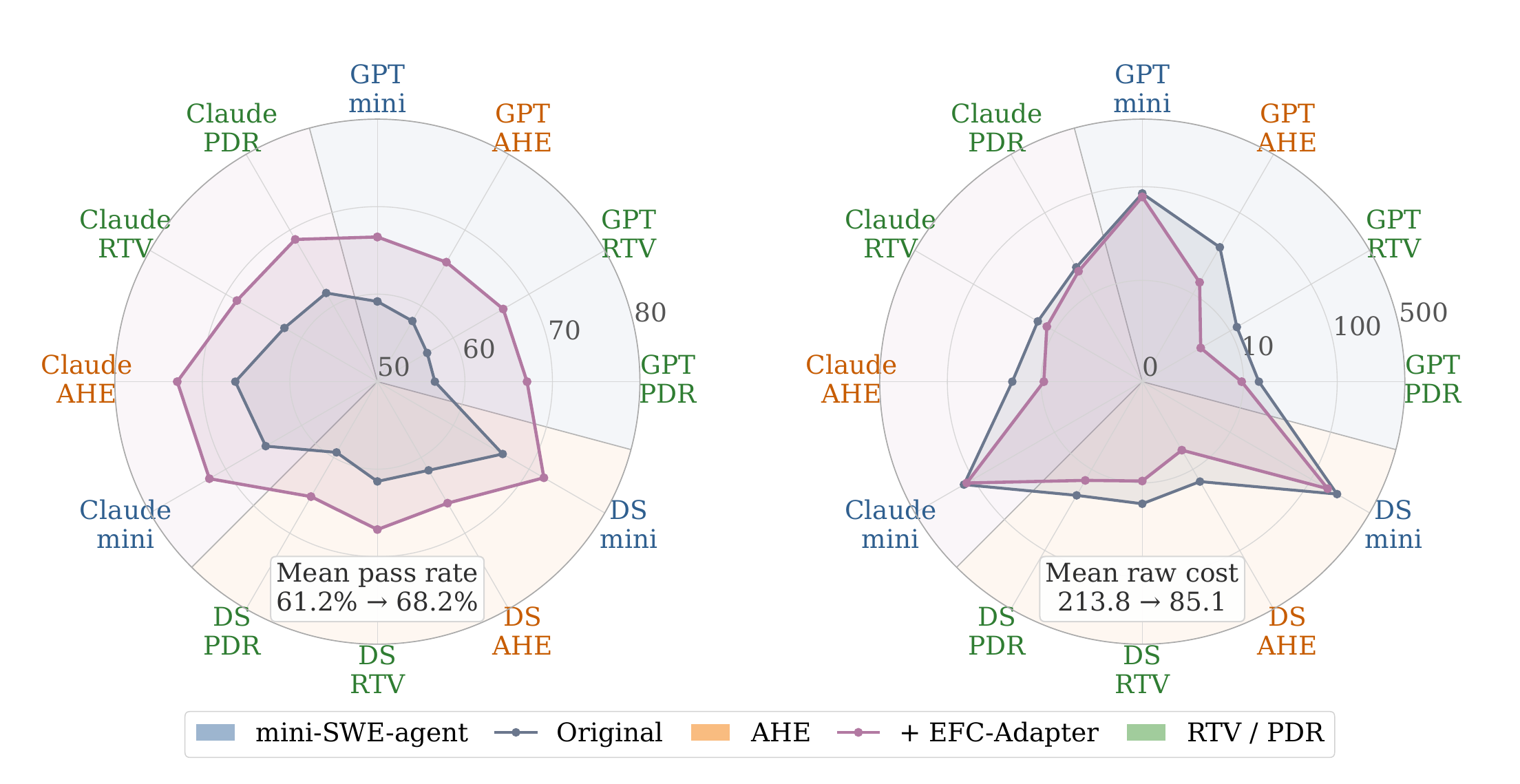}
    \caption{
    \textbf{\ours improves pass rate and reduce raw cost.}
    Each spoke is a model--method pair across gpt-5.4-nano (GPT), DeepSeek-V4-Flash (DS), and Claude-Haiku-4.5 (Claude) with mini-SWE-agent (mini), AHE, RTV, or PDR.
    The two radars report pass rate and raw cost.
    }
    \label{fig:efc_adapter_radar}
\end{figure}

As shown in Figure~\ref{fig:efc_adapter_radar}, \ours improves the accuracy--cost tradeoff.
It raises the overall mean pass rate from $61.2\%$ to $68.2\%$ while reducing
mean raw cost (Eq.~\ref{eq:raw_cost_appendix}) from $213.8$ to $85.1$, under the
same model, task split, tool interface, action space, and raw-budget caps.
\emph{(i)} The gains vary with task structure: HumanEval benefits from clean and localized execution signals, Terminal-Bench benefits from filtering noisy or repetitive shell feedback, and SWE-bench shows smaller but steady gains because repository-level failures often require longer-range context.
\emph{(ii)} RTV and PDR benefit from selecting among candidate feedback traces with sharply different value, and mini-SWE-agent and AHE mainly benefit from suppressing low-value repair or update loops.
\emph{(iii)} Across models, the adapter helps weaker models most, suggesting that our method can partially compensate for limited self-diagnosis, and still reduce redundant or misleading feedback for stronger models.

\section{Conclusion}
This paper shows that agent harness scaling is better explained by effective
feedback than by raw test-time expenditure.
We introduce Effective Feedback Compute (EFC), with Estimated-EFC, NRS-EFC,
harness efficiency, and task-demand normalization, to measure valid,
informative, non-redundant, and retained feedback relative to task demand.
Across controlled, real, and held-out evaluations,
EFC-based coordinates outperform raw-compute baselines and SAS, and \ours
improves the accuracy--cost tradeoff of existing harnesses under matched
settings.
Overall, harness scaling is best viewed as converting raw interaction into
durable, task-sufficient feedback.

\clearpage

\bibliography{tacl2021}
\bibliographystyle{acl_natbib}

\clearpage

\appendix
\section{Related Work}
    \label{sec:related}
    \paragraph{Scaling and test-time compute.}
Scaling laws connect language-model performance with model size, data size, and
training compute \citep{kaplan2020scaling,hoffmann2022training}. Recent work
extends this view to inference-time computation, showing that repeated sampling,
search, revision, and adaptive allocation can improve performance when additional
test-time budget is available
\citep{brown2024large,snell2025scaling,zhu2025scaling_agents,kim2026agenticcoding}.
These studies establish test-time compute as an important scaling axis, but they
usually measure compute through samples, tokens, rollouts, or search budget. Our
work instead asks which scalar predicts when this extra computation becomes
useful. EFC answers this by counting feedback that is informative, valid,
non-redundant, and retained, then normalizing it by task demand.

\paragraph{Harness feedback and agent-system scaling.}
Agent harnesses improve language models by adding reasoning, action, verification, search, memory, and iterative feedback \citep{yao2023react,shinn2023reflexion,madaan2023selfrefine,yao2023tree,zhou2024lats,lightman2023verify}. Recent work increasingly treats the harness itself as a first-class object of design and evaluation, including natural-language harness representations, versioned agent-optimization loops, and context-retrieval benchmarks for coding agents \citep{pan2026natural_harnesses,ursekar2026vero,li2026contextbench}. System-level studies further argue that agent performance depends on the joint scaling of model capability, orchestration, verification, coordination, and overhead rather than on base-model scale alone \citep{kim2026sas,li2026general_agentbench}; similarly, skill-library scaling results show that routing and execution behavior depend on library size, granularity, and exposure policy \citep{chen2026skillscaling}. These works motivate harness-level scaling, while our focus is complementary: we seek a trace-level scalar coordinate that separates raw expenditure from useful retained feedback. EFC therefore provides a compact predictive quantity for closed-loop harness scaling and is directly compared with SAS as a strong system-level baseline.

\section{Task Details}
\label{app:tasks}

This appendix describes the task layers used in our experiments. The purpose of
the task suite is not to maximize benchmark coverage, but to span a controlled
range from oracle-observable feedback to realistic agent trajectories.

\subsection{Synthetic Controllable Tasks}
\label{app:synthetic_tasks}

The synthetic layer contains procedurally generated tasks with hidden state and
known ground-truth solution paths. Each task instance specifies a target state,
a set of candidate hypotheses, a tool interface, and a deterministic evaluator.
Because the latent state is known to the experimenter, we can compute
Oracle-EFC for each feedback event.

We use this layer for three purposes. First, it provides controlled variation in
task demand, including required solution length, tool-selection entropy,
state size, observation noise, and verifier availability. Second, it allows us to
test whether feedback events measured by EFC correspond to real progress toward
the solution. Third, it provides calibration data for Estimated-EFC, which is
later applied to tasks where oracle state is unavailable.

Each synthetic task is generated from a template family. Template parameters
control the number of latent variables, the number of distractor tools, the
probability of noisy observations, and the coverage of deterministic checkers.
We split generated tasks into calibration and held-out sets by template family
and random seed.

\subsection{Semi-Realistic Executable Tasks}
\label{app:semi_real_tasks}

The semi-realistic layer contains tasks with realistic artifacts and executable
checkers. Examples include small code-repair tasks, data-analysis tasks, and
tool-use tasks with deterministic validation. These tasks are designed to produce
natural agent traces while preserving reliable evaluation.

For code tasks, the final answer is executed against unit tests or reference
checks. Intermediate checker results, such as syntax errors, failing tests, and
partial correctness signals, are logged as trace observations. For tool-use
tasks, the environment exposes a fixed set of tools and records whether tool
outputs are later referenced or used in solution updates.

This layer is used to evaluate whether the EFC estimators calibrated on
synthetic tasks remain predictive when traces contain realistic model errors,
ambiguous observations, and incomplete verification.

\subsection{Real Benchmarks}
\label{app:real_benchmarks}

The real benchmark layer uses verifiable subsets of existing agent benchmarks.
We include only tasks for which final success can be evaluated automatically and
for which traces can be logged in a consistent format. The benchmark layer is
used for external validity rather than leaderboard comparison.

We apply the following filtering rules:
\begin{itemize}
    \item the task must have an automatic final evaluator;
    \item the required environment must be reproducible in a local or sandboxed
    execution setting;
    \item the task must admit meaningful intermediate observations, such as test
    results, command outputs, or tool responses;
    \item the task must fit within the budget limits used by the harness family.
\end{itemize}

When a benchmark contains tasks with very large setup cost or unstable external
dependencies, we exclude those tasks from the main analysis and report the
filtering rule separately. The resulting subset is intended to test scaling-law
generalization, not to estimate absolute benchmark performance.

\subsection{Task Demand Variables}
\label{app:task_difficulty}

For each task, we compute a hand-designed task-demand score following Eq.~\ref{eq:dtask}.
The terms are defined as follows:
\begin{itemize}
    \item $L$ is the estimated minimum number of reasoning or action steps.
    \item $H_{\mathrm{tool}}$ measures tool-selection entropy or the number of
    plausible but incorrect tools.
    \item $S_{\mathrm{state}}$ measures the amount of state that must be tracked
    across the trajectory.
    \item $N_{\mathrm{obs}}$ measures observation noise, ambiguity, or
    nondeterminism.
    \item $V_{\mathrm{ver}}$ measures verifier-signal visibility, namely the
    availability and coverage of reliable task-level validation signals such as
    deterministic checkers, explicit tests, partial evaluators, or benchmark
    scoring hooks.
\end{itemize}

In real benchmark traces, $V_{\mathrm{ver}}$ is computed from task metadata
about verification coverage rather than from the agent's final outcome or a
hidden solution. The agent does not observe oracle answers through this
quantity. The factor appears as $1-V_{\mathrm{ver}}$ because tasks with
stronger reliable verification signals require less feedback mass to reach the
same effective progress.

All components are normalized using statistics estimated on the calibration
split within each task layer. The resulting normalization constants are frozen
before held-out and prospective evaluation. In addition to this hand-designed
score, we also evaluate fitted task-demand weights learned on calibration tasks
and applied to held-out tasks.

\subsection{Run Logging}
\label{app:run_logging}

For every run, we log the task identifier, task family, harness identifier,
model identifier, budget configuration, final success label, raw compute
variables, checker outputs, and the full sequence of trajectory events. Each
event records the action type, observation type, tool name if applicable,
checker result if available, memory update if present, and references to earlier
observations. These logs are the sole input to Estimated-EFC on non-oracle
tasks.

\section{EFC Factor Measurement}
\label{app:efc_factor_measurement}

This appendix specifies how the event factors $I_t,V_t,R_t,M_t$ are computed or
estimated in each task layer. Section~\ref{sec:efc} defines the common EFC
variables and aggregation rules.

\subsection{Common Notation}
\label{app:efc_factor_common}

We use $\operatorname{clip}(x)$ to denote clipping to $[0,1]$:
\begin{equation}
    \operatorname{clip}(x)
    =
    \min
    \{
    1,
    \max
    \{
    0,
    x
    \}
    \}.
    \label{eq:clip}
\end{equation}
For every event, the factors are computed before aggregation. The run-level
value is then obtained by Eq.~\ref{eq:efc}.

When a feature is unavailable in a task layer, we use the closest trace-observable
proxy. The final answer correctness label is excluded from all event-level
factor estimates.
All scalar constants used in the factor proxies, including $\kappa$, $\alpha_A$,
status-quality weights, progress-gate weights, and loop-gate weights, are fixed
on the calibration split before held-out evaluation. We do not tune these
constants on held-out task families, real benchmark subsets, or the prospective
batch.

\subsection{Synthetic Controllable Tasks}
\label{app:efc_factor_synthetic}

In the synthetic layer, the experimenter observes the latent task state and the
ground-truth solution path. Oracle-EFC therefore uses direct event-level factors.
Let $n_t$ denote novelty relative to previous events, $B^{\mathrm{noise}}_t$
the reliability of the observation channel, $B^{\mathrm{route}}_t$ routing
quality, $B^{\mathrm{verify}}_t$ verifier strength, and
$B^{\mathrm{mem}}_t$ memory fidelity. We instantiate the factors as
\begin{equation}
    I_t
    =
    \operatorname{clip}
    \left(
    B^{\mathrm{route}}_t
    B^{\mathrm{noise}}_t
    \Delta^{\mathrm{latent}}_t
    \right),
    \label{eq:synthetic_i}
\end{equation}
\begin{equation}
    V_t
    =
    \operatorname{clip}
    \left(
    B^{\mathrm{verify}}_t
    B^{\mathrm{noise}}_t
    V_{\mathrm{er}}
    \right),
    \label{eq:synthetic_v}
\end{equation}
\begin{equation}
    R_t
    =
    \operatorname{clip}
    \left(
    n_t
    B^{\mathrm{route}}_t
    B^{\mathrm{tool}}_t
    \right),
    \label{eq:synthetic_r}
\end{equation}
\begin{equation}
    M_t
    =
    \operatorname{clip}
    \left(
    B^{\mathrm{mem}}_t
    B^{\mathrm{state}}_t
    (0.82 + 0.18 V_t)
    \right).
    \label{eq:synthetic_m}
\end{equation}
Here $\Delta^{\mathrm{latent}}_t$ is ground-truth progress toward the target
state, $B^{\mathrm{tool}}_t$ discounts high tool ambiguity, and
$B^{\mathrm{state}}_t$ discounts memory pressure in large state spaces.

For the three synthetic task families, $\Delta^{\mathrm{latent}}_t$ is defined
as follows:
\begin{itemize}[leftmargin=*]
    \item \textbf{Needle Lookup.}
    Let $C_t$ be the remaining candidate set and let $b_t$ indicate whether the
    target key-value relation is recovered. We use
    $\Delta^{\mathrm{latent}}_t
    =
    \operatorname{clip}
    ((|C_{t-1}|-|C_t|)/\max(1,|C_{t-1}|) + b_t)$.

    \item \textbf{State Tracking.}
    Let $d_t$ be the number of correct state transitions committed after event
    $t$ and let $r^{\mathrm{fix}}_t$ indicate correction of a previous state
    error. We use
    $\Delta^{\mathrm{latent}}_t
    =
    \operatorname{clip}
    ((d_t-d_{t-1})/\max(1,L) + r^{\mathrm{fix}}_t)$.

    \item \textbf{Rule Filter.}
    Let $E_t$ be the number of eliminated nonmatching items and $P_t$ the number
    of confirmed matching conditions. We use
    $\Delta^{\mathrm{latent}}_t
    =
    \operatorname{clip}
    ((E_t-E_{t-1}+P_t-P_{t-1})/\max(1,N_{\mathrm{items}}))$.
\end{itemize}

The Oracle-EFC event score is
\begin{equation}
    \mathrm{EFC}^{\mathrm{oracle}}_t
    =
    \kappa I_t V_t R_t M_t.
    \label{eq:synthetic_oracle_event}
\end{equation}

\subsection{Semi-Realistic Executable Tasks}
\label{app:efc_factor_semireal}

In the semi-realistic layer, traces contain executable feedback and realistic
model errors. We compute factor estimates from observable event features. Let
$c_t$ be checker fired, $h_t$ checker scope, $z_t$ tool-result reference,
$p_t$ plan update, $m_t$ memory retention, $a_t$ repeated-error avoidance,
$q_t$ observation consistency, $\Delta_t$ subgoal progress, and $n_t$ novelty.
Let $B_{\mathrm{router}}$ be harness routing quality, $B_{\mathrm{verify}}$
verifier strength, $H_{\mathrm{tool}}$ tool ambiguity, and
$E_{\mathrm{explore}}$ exploration entropy.

We estimate informativeness as
\begin{equation}
    \widehat{I}_t
    =
    \operatorname{clip}
    \left(
    \frac{
    \Delta_t
    (0.70 + 0.30 B_{\mathrm{router}})
    }{
    1 + 0.12 H_{\mathrm{tool}} + 0.20 E_{\mathrm{explore}}
    }
    \right).
    \label{eq:semireal_i}
\end{equation}
We estimate validity as
\begin{equation}
    \widehat{V}_t
    =
    \operatorname{clip}
    \left(
    q_t
    (0.70 + 0.30 B_{\mathrm{verify}})
    (0.72 + 0.28 V_{\mathrm{ver}})
    \right).
    \label{eq:semireal_v}
\end{equation}
We estimate non-redundant relevance as
\begin{equation}
    \widehat{R}_t
    =
    \operatorname{clip}
    \left(
    \frac{
    (0.28 + 0.72 n_t)
    (0.48 + 0.52 a_t)
    }{
    1 + 0.12 H_{\mathrm{tool}}
    }
    \right).
    \label{eq:semireal_r}
\end{equation}
We estimate memory update as
\begin{equation}
    \widehat{M}_t
    =
    \operatorname{clip}
    \left(
    m_t(0.80 + 0.20 p_t)
    \right).
    \label{eq:semireal_m}
\end{equation}

For calibration traces with oracle-visible progress, these factor estimates
define the event target
\begin{equation}
    y_t
    =
    \kappa
    \widehat{I}_t
    \widehat{V}_t
    \widehat{R}_t
    \widehat{M}_t.
    \label{eq:semireal_target}
\end{equation}
The trace estimator in Eq.~\ref{eq:estimated_event_efc} is trained to predict
$y_t$ from the feature vector in Eq.~\ref{eq:efc_trace_features}.

\subsection{Real Benchmarks}
\label{app:efc_factor_real}

For HumanEval, Terminal-Bench 2.0, and SWE-bench Verified,  hidden state is unavailable. We first compute
$\widehat{\mathrm{EFC}}_t$ with the calibrated trace estimator. We then apply
deterministic status and repetition gates derived from execution traces.

The status-quality factor $Q_t$ is
\begin{equation}
    Q_t
    =
    \begin{cases}
    1.00, & \text{passed},\\
    0.42, & \text{assertion error},\\
    0.12, & \text{runtime error},\\
    0.06, & \text{timeout},\\
    0.04, & \text{static reject or missing entry point},\\
    0.00, & \text{API error},\\
    0.25, & \text{other status}.
    \end{cases}
    \label{eq:real_status_quality}
\end{equation}
Let $\operatorname{sev}(s_t)$ map event status to ordered severity, with
API errors lowest and passing checks highest. The progress gate is
\begin{equation}
    G_t
    =
    \begin{cases}
    1.00, & A_t = 0,\\
    1.35, & s_t=\text{passed} \text{ and } s_{t-1}\neq\text{passed},\\
    1.15, & \operatorname{sev}(s_t)>\operatorname{sev}(s_{t-1}),\\
    0.62, & \operatorname{sev}(s_t)=\operatorname{sev}(s_{t-1})
    \text{ and } s_t\neq\text{passed},\\
    0.45, & \operatorname{sev}(s_t)<\operatorname{sev}(s_{t-1}),\\
    1.00, & \text{otherwise}.
    \end{cases}
    \label{eq:real_progress_gate}
\end{equation}
The loop gate is
\begin{equation}
    \Lambda_t
    =
    \begin{cases}
    0.95, & \text{repair event},\\
    0.92, & \text{generation event with passing status},\\
    0.85, & \text{generation event without passing status},\\
    1.00, & \text{otherwise}.
    \end{cases}
    \label{eq:real_loop_gate}
\end{equation}

For nonredundant stable EFC, repeated failures receive stronger discounts:
\begin{equation}
    G^{\mathrm{nr}}_t
    =
    \begin{cases}
    1.00, & A_t = 0,\\
    1.35, & s_t=\text{passed} \text{ and } s_{t-1}\neq\text{passed},\\
    1.15, & \operatorname{sev}(s_t)>\operatorname{sev}(s_{t-1}),\\
    0.16, & \operatorname{sev}(s_t)=\operatorname{sev}(s_{t-1})
    \text{ and } s_t\neq\text{passed},\\
    0.10, & \operatorname{sev}(s_t)<\operatorname{sev}(s_{t-1}),\\
    1.00, & \text{otherwise}.
    \end{cases}
    \label{eq:real_nr_progress_gate}
\end{equation}
The nonredundant loop gate is
\begin{equation}
    \Lambda^{\mathrm{nr}}_t
    =
    \begin{cases}
    0.45, & \text{repair event},\\
    0.92, & \text{generation event with passing status},\\
    0.85, & \text{generation event without passing status},\\
    1.00, & \text{otherwise}.
    \end{cases}
    \label{eq:real_nr_loop_gate}
\end{equation}

These gates correspond to the real-trace factor proxies
$\widehat{V}^{\mathrm{real}}_t=Q_t$ and
$\widehat{R}^{\mathrm{real}}_t=G_t\Lambda_t$. The base estimator supplies the
trace-level product of informativeness, validity, relevance, and memory
retention from Eq.~\ref{eq:estimated_event_efc}. The status-aware scores used in
the real benchmark analyses are those in Eqs.~\ref{eq:stable_estimated_efc} and \ref{eq:nonredundant_stable_efc}.

\subsection{Raw-Cost Accounting}
\label{app:raw_cost}

For each trajectory $\tau$, we record four trace-observable resource variables:
the total number of model tokens $N_{\mathrm{tok}}(\tau)$, the number of external
tool or checker calls $N_{\mathrm{tool}}(\tau)$, the elapsed wall-clock time
$T_{\mathrm{wall}}(\tau)$, and the number of harness operations
$N_{\mathrm{op}}(\tau)$. The realized raw-compute cost is computed as
\begin{equation}
\begin{aligned}
C_{\mathrm{raw}}(\tau)
=
\lambda_{\mathrm{tok}}
\frac{N_{\mathrm{tok}}(\tau)}{U_{\mathrm{tok}}}
+
\lambda_{\mathrm{tool}} N_{\mathrm{tool}}(\tau)\\
+
\lambda_{\mathrm{time}} T_{\mathrm{wall}}(\tau)
+
\lambda_{\mathrm{op}} N_{\mathrm{op}}(\tau).
\label{eq:raw_cost_appendix}
\end{aligned}
\end{equation}
In all experiments, we use $U_{\mathrm{tok}}=1000$ tokens,
$\lambda_{\mathrm{tok}}=1.0$, $\lambda_{\mathrm{tool}}=0.35$,
$\lambda_{\mathrm{time}}=0.10$, and $\lambda_{\mathrm{op}}=0.04$. These weights
define a normalized accounting unit for comparing raw expenditure across runs.

\section{Detailed Experimental Setup}
\label{app:detailed_experimental_setup}

\subsection{Task Layers}
\label{sec:tasks}

We evaluate harness scaling on three task layers that progressively reduce
oracle access while preserving automatic evaluation.

\begin{itemize}
    \item \textbf{Synthetic controllable tasks.}
    We use procedurally generated Needle Lookup, State Tracking, and Rule Filter
    tasks with hidden state and deterministic answers. This layer supports
    direct measurement of Oracle-EFC and controlled variation in
    $D_{\mathrm{task}}$.

    \item \textbf{Semi-realistic executable tasks.}
    We use code tasks and small executable repair or analysis
    tasks with unit-test or reference-check feedback. This layer tests whether
    Estimated-EFC remains predictive when traces contain realistic model errors.

    \item \textbf{Real benchmark subsets.}
    We use verifiable HumanEval~\citep{chen2021humaneval}, Terminal-Bench 2.0 (Terminal-Bench)~\citep{merrill2026terminalbench}, and SWE-bench Verified (SWE-bench)~\citep{jimenez2024swebench}.
    This layer tests transfer to realistic agent trajectories.
\end{itemize}


Unless otherwise stated, run-level quantities are first computed per run and then aggregated into the evaluation groups used by each experiment.

\subsection{Harness Families}
\label{sec:harnesses}

We compare seven harness families, denoted H0--H6, that differ in how they
convert raw budget into useful feedback.

\begin{enumerate}
    \item[\textbf{H0}] \textbf{Direct Answer.}
    The model produces a solution in one pass, without explicit tool feedback,
    verification, repair, or memory.

    \item[\textbf{H1}] \textbf{Checklist Verify.}
    The harness adds lightweight verification or checklist-style checks to the
    direct solution process, providing limited feedback without a full
    closed-loop controller.

    \item[\textbf{H2}] \textbf{Routed Tools.}
    The harness routes between available tools and model calls, allowing the
    agent to condition later actions on external observations.

    \item[\textbf{H3}] \textbf{Stateful Memory.}
    The harness maintains compact memory over verified facts, failed attempts,
    and task constraints, reducing repeated errors across steps.

    \item[\textbf{H4}] \textbf{High Budget Noisy.}
    The harness spends a larger raw budget under weaker routing, verification,
    and memory conditions, isolating raw expenditure from effective feedback.

    \item[\textbf{H5}] \textbf{Closed Loop.}
    The harness combines routing, verification, and structured memory in an iterative loop, allowing observations to be checked, retained, and reused in subsequent decisions.

    \item[\textbf{H6}] \textbf{Deep Closed Loop.}
    The harness extends the closed-loop setting with a larger interaction depth
    and stronger feedback mechanisms, testing whether additional budget helps
    when it is converted into effective feedback.
\end{enumerate}





\subsection{Harness Details}
\label{app:harnesses}

This appendix provides additional details on the harness families used in our
experiments. The seven harnesses correspond exactly to H0--H6 in
Section~\ref{sec:harnesses}. They are designed to vary how raw computation is
converted into effective feedback, while keeping the task distribution, base
model, final evaluator, and logging protocol fixed within each experimental
setting.

\subsubsection{Common Harness Interface}

Each run is specified by a task instance, a base model, a harness family, and a
replicate index. The harness produces a final answer and a trajectory of
intermediate events. These events record the interaction structure of the run,
including whether the model receives external observations, verifier signals,
repair feedback, routed information, or retained state from previous steps. The
same trace schema is used across controlled, semi-realistic, and real benchmark
settings, enabling all harnesses to be compared through the same EFC-based
metrics.

The harnesses differ along six main dimensions: raw budget, tool budget,
verifier strength, routing quality, memory fidelity, and noise/state pressure.
Raw budget measures the amount of computation made available to the harness,
whereas the other dimensions determine how efficiently this computation is
converted into useful, valid, remembered, and non-redundant feedback. This
separation is important because a harness can spend substantially more raw
budget without necessarily producing proportionally more EFC.

\subsubsection{H0: Direct Answer}

H0 is the direct-answer baseline. The model receives the task instruction and
produces a final answer in a single pass. It does not use routed observations,
explicit verification, persistent memory, or feedback-conditioned repair before
submission. H0 therefore measures the performance obtainable from the base model
with minimal harnessing.

In the EFC interpretation, H0 has low expected feedback mass: most computation
is spent on direct generation rather than on acquiring, validating, or reusing
external feedback. It serves as the lower-complexity reference point for
measuring the effect of increasingly structured harness mechanisms.

\subsubsection{H1: Checklist Verify}

H1 augments direct generation with a lightweight verification or checklist
stage. The model is encouraged to check constraints, edge cases, or internal
consistency before finalizing its answer. This mechanism increases the amount of
self- or verifier-guided scrutiny relative to H0, but it does not create a full
closed loop: the run remains primarily single-pass and does not rely on
multi-round repair.

H1 isolates the effect of adding a weak verification signal without introducing
strong routing, persistent memory, or deep interaction. It is therefore useful
for distinguishing simple checking from the stronger feedback accumulation used
by H5 and H6.

\subsubsection{H2: Routed Tools}

H2 introduces routed tool or observation access. Rather than treating all
available information uniformly, the harness allocates part of its budget to
selecting which observations are likely to be useful for the current task state.
This improves the relevance of feedback events relative to H0/H1, especially
when the task benefits from external evidence or intermediate computation.

However, H2 does not assume strong memory or deep verifier-driven repair. Its
main mechanism is improved routing: the harness can obtain more informative
observations, but it has only moderate ability to validate, retain, and
iteratively refine them. H2 therefore tests whether better access to observations
alone is sufficient to explain the observed scaling behavior.

\subsubsection{H3: Stateful Memory}

H3 emphasizes state retention. Compared with H2, it gives the harness a stronger
ability to preserve useful intermediate information, remember failed attempts,
and avoid repeating previously identified errors. The intended mechanism is not
simply to spend more budget, but to make prior feedback more reusable across
later steps.

In trace terms, H3 increases the probability that valid feedback remains
available and affects subsequent decisions. This makes it a natural test case
for the memory component of EFC: feedback that is observed but forgotten has
limited effect, whereas feedback that is retained can continue to shape future
actions.

\subsubsection{H4: High Budget Noisy}

H4 is a high-budget but inefficient harness. It allocates substantially more raw
computation than the simpler harnesses, but combines this budget with weaker
routing, weaker verification, weaker memory, higher observation noise, and
greater state pressure. As a result, H4 may generate many intermediate events
without converting them into proportionally useful feedback.

This harness serves as a negative control for raw-budget explanations. If
performance were primarily determined by tokens, tool calls, wall-clock time, or
other raw compute proxies, H4 should be highly competitive. Instead, its role is
to test whether additional computation only helps when the harness can transform
that computation into valid and non-redundant feedback.

\subsubsection{H5: Closed Loop}

H5 is the standard closed-loop harness. It combines stronger routing,
verification, and memory with feedback-conditioned refinement. The harness can
use intermediate observations or verifier signals to revise its subsequent
behavior, rather than treating each attempt as independent. This creates a
structured feedback loop in which the model proposes, receives task-relevant
signals, and updates its next action accordingly.

Compared with H2 and H3, H5 is not defined by a single mechanism such as routing
or memory alone. Its advantage comes from the interaction among mechanisms:
routing increases the relevance of observations, verification improves their
validity, and memory allows useful information to persist across the loop. This
makes H5 the first harness family expected to achieve consistently high EFC
efficiency.

\subsubsection{H6: Deep Closed Loop}

H6 extends H5 by increasing interaction depth and strengthening the feedback
conversion mechanisms. It uses a deeper closed-loop process with stronger
verification, better routing, higher memory fidelity, and lower effective
observation noise. The additional budget in H6 is therefore not merely more raw
computation; it is budget deployed through a harness that is better able to turn
intermediate signals into effective feedback.

H6 represents the strongest harness family in our experimental design. It tests
whether scaling continues when a harness already has high feedback efficiency,
and whether deeper interaction provides additional gains beyond the standard
closed-loop structure of H5. In the EFC framework, H6 is expected to produce not
only more feedback events, but also feedback events with higher validity,
relevance, retention, and non-redundancy.

\subsubsection{Comparison Across Harness Families}

The harness sequence H0--H6 is not intended to be a simple monotonic increase in
raw compute. Instead, it separates raw budget from feedback efficiency. H0 and
H1 provide low-interaction baselines; H2 and H3 isolate routing and memory; H4
provides a high-budget but noisy counterexample; and H5/H6 instantiate
increasingly strong closed-loop agents. This design allows us to evaluate
whether performance is better explained by raw computation or by EFC.

Across experiments, the same harness names refer to the same mechanism-level
roles. Controlled experiments instantiate these roles through explicit
simulation parameters, while semi-realistic and real benchmark experiments
instantiate them through executable harness behavior and trace-observable
feedback events. This shared abstraction makes it possible to compare harnesses
across settings without relying on benchmark-specific implementation details.

\subsubsection{Stopping Rules}

A run terminates when the harness submits a final answer, when the configured
interaction budget is exhausted, when a verifier or evaluator accepts the
current candidate, or when the environment returns an unrecoverable execution
failure. Budget exhaustion is counted as failure unless the final submitted
answer passes the evaluator. These stopping rules are applied consistently
within each experimental setting so that differences across H0--H6 reflect
harness structure rather than evaluation protocol.

\subsection{Scalar Predictors and Baselines}
\label{sec:raw_baselines}
\label{sec:scalar_predictors}

We compare EFC-based predictors against standard raw-compute baselines and a
strong system-level baseline. For each run, we record the following
trace-observable quantities:

\begin{itemize}
    \item \textbf{Raw Tokens}: total input and output tokens consumed by the model.
    \item \textbf{Tool calls}: number of external tool invocations.
    \item \textbf{Wall time}: elapsed runtime of the harness.
    \item \textbf{Operations}: total number of model, tool, verification,
    and memory-update operations.
    \item \textbf{Raw cost}: a normalized cost combining token usage, tool calls,
    operations, and runtime (Eq.~\ref{eq:raw_cost_appendix}).
\end{itemize}

We also include \textbf{SAS}~\citep{kim2026sas}, a prior agent-systems scaling baseline that uses a fixed-effect equation over system-level quantities, and implement it following the original equation and implementation details. 
SAS serves as a strong multivariate baseline for system-level scaling models.


\section{Additional Experiments}
\label{app:additional_experiments}

\subsection{Executable Code Tasks Preserve the EFC Signal}
\label{sec:semireal_transfer}

We next test whether the trace-time EFC signal survives in a more realistic
code-execution setting. We use held-out executable programming tasks in a function-completion format, where candidate programs are checked
against executable tests and repaired over multiple attempts. The harness
observes syntax errors, runtime errors, unit-test failures, partial correctness
signals, and subsequent repair behavior. Estimated-EFC is computed only from
these trace-time signals, while oracle quantities are used only for diagnostic
comparison. 
We also evaluate seven harness variants and report estimated feedback efficiency, measuring how much Estimated-EFC is produced per unit of raw cost.

\begin{figure*}[t]
    \centering
    \includegraphics[width=.9\linewidth]{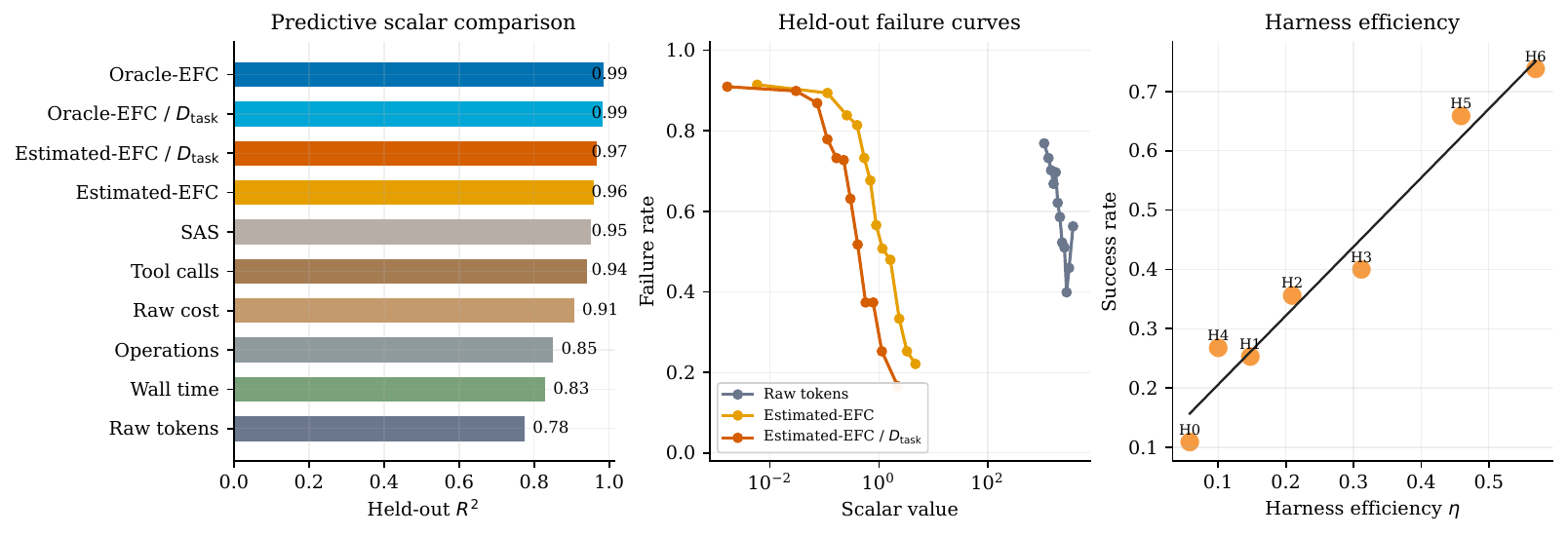}
    \caption{
    \textbf{Executable code tasks preserve the EFC signal.}
    Left: on held-out programming tasks, EFC-based predictors achieve the highest $R^2$. 
    Middle: failure curves for Estimated-EFC and Estimated-EFC/$D_{\mathrm{task}}$ are cleaner than the raw-token curve, indicating that execution feedback quality is more informative than generation length. 
    Right: harness success increases with $\eta$, showing that harness variants succeed when they convert raw budget into effective feedback more efficiently.
    }
    \label{fig:executable_code}
\end{figure*}

Figure~\ref{fig:executable_code} shows that the EFC signal persists in
executable code tasks. \emph{(i)} Tool calls become a strong raw baseline when
they partially align with executable feedback. Since tool calls often run tests
that expose syntax errors, runtime errors, or unit-test failures, they reach
$R^2=0.94$, far above raw tokens at $0.78$ and close to SAS at $0.95$.
\emph{(ii)} EFC-based coordinates remain stronger because they score what the
tests reveal and whether the harness uses that information. Estimated-EFC
reaches $0.96$, Estimated-EFC/$D_{\mathrm{task}}$ reaches $0.97$, and oracle
variants reach $0.99$. \emph{(iii)} Feedback efficiency explains harness-level
differences. Shallow or noisy harnesses remain low-success even when they spend
budget, while H5 and H6 achieve the highest success by converting raw cost into
higher estimated feedback efficiency. 
Thus, executable feedback strengthens the interpretation that tool use helps when it produces useful, retained evidence, not merely because it increases interaction count.

\subsection{Module Ablations Localize Raw-to-EFC Conversion Gains}
\label{sec:module_ablation}

We next localize the raw-to-EFC conversion gains to concrete harness components.
We ablate three harness modules and one interface corruption factor: verifier
strength, memory fidelity, router quality, and observation noise. Each family is
evaluated at ordered settings, ranging from disabled or weak configurations to
stronger configurations that provide more reliable feedback. We measure both
harness efficiency $\eta$ and downstream success.

\begin{figure*}[t]
    \centering
    \includegraphics[width=.9\linewidth]{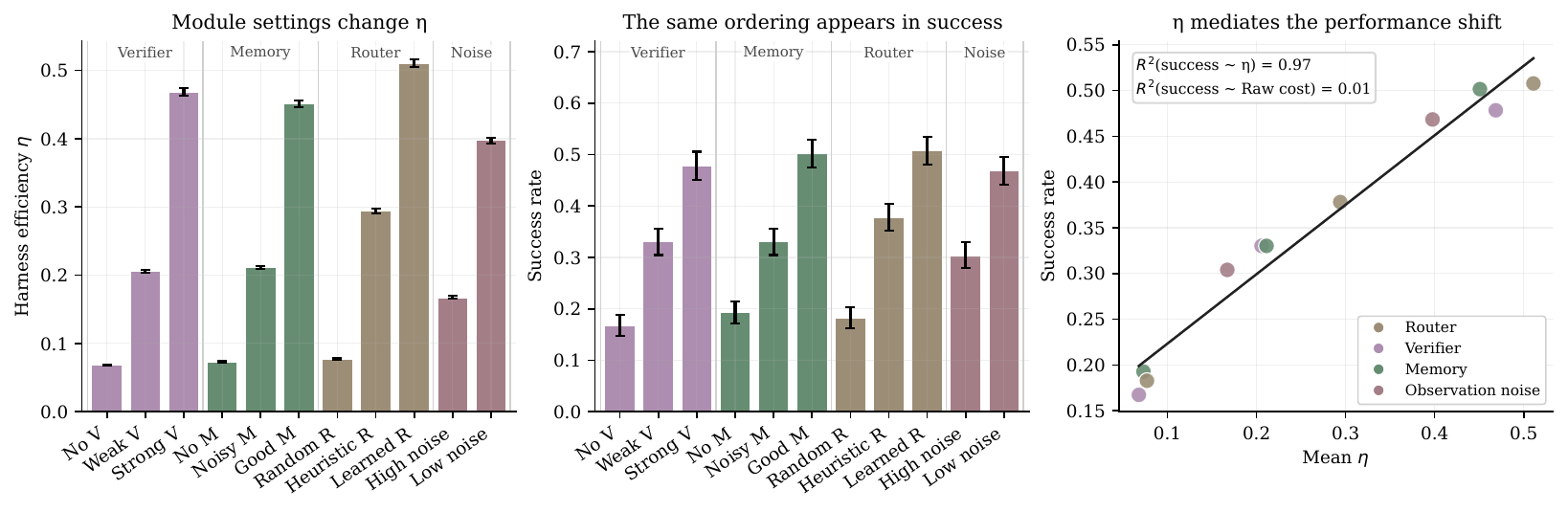}
    \caption{
    \textbf{Module ablations localize raw-to-EFC conversion gains.}
    Left: stronger verifier, memory, and router modules increase harness
    efficiency $\eta$, while lower observation noise also improves $\eta$.
    Middle: the same ordering appears in success rate, showing that module
    improvements that increase $\eta$ also improve task outcomes. 
    Vertical error bars in the two bar panels denote approximate 95\% confidence intervals ($1.96$ SEM), and thin gray separators group settings by module family.
    Vertical error bars show mean $\pm 1.96$ standard errors across repeated runs.
    Right: success is almost fully explained by mean $\eta$
    ($R^2=0.97$), whereas raw cost has essentially no explanatory power
    ($R^2=0.01$), indicating that $\eta$ mediates the performance shift.
    }
    \label{fig:eta_mechanism}
\end{figure*}

Figure~\ref{fig:eta_mechanism} shows that raw-to-EFC conversion can be localized to specific harness modules. 
\emph{(i)} The ablations produce a dose-response pattern. 
Moving from disabled or weak modules to stronger verifier, memory, and router settings increases both $\eta$ and success, while reducing observation noise produces the same direction of improvement. 
Router quality reaches the highest mean $\eta$ and success among the module settings, matching the factor scan in Figure~\ref{fig:eta_decomposition}. 
\emph{(ii)} Harness efficiency mediates the performance shift, while raw cost does not. 
Across module settings, mean $\eta$ explains almost all variation in success ($R^2=0.97$), whereas raw cost explains almost none ($R^2=0.01$). 
This localizes the mechanism: modules help when they improve validity, retention, relevance, non-redundancy, or observation quality, not when they merely change expenditure.

\subsection{Model-Specific Results for Main Experiments}
\label{app:model_specific_results}

The main text reports results averaged across \texttt{DeepSeek-V4-Flash},
\texttt{gpt-5.4-nano}, and \texttt{Claude-Haiku-4.5}. 
Here we repeat the main analyses separately for each base model, using the same
grouping, fitting, and evaluation protocol as in the corresponding main-text
experiments. 
These results test whether the reported EFC trends are robust across base-model
capability levels rather than being driven by a single backbone.

\subsubsection{Model-Specific Controlled Scaling Results}
\label{app:model_specific_controlled_scaling}

\begin{figure*}[t]
    \centering
    \includegraphics[width=.95\textwidth]{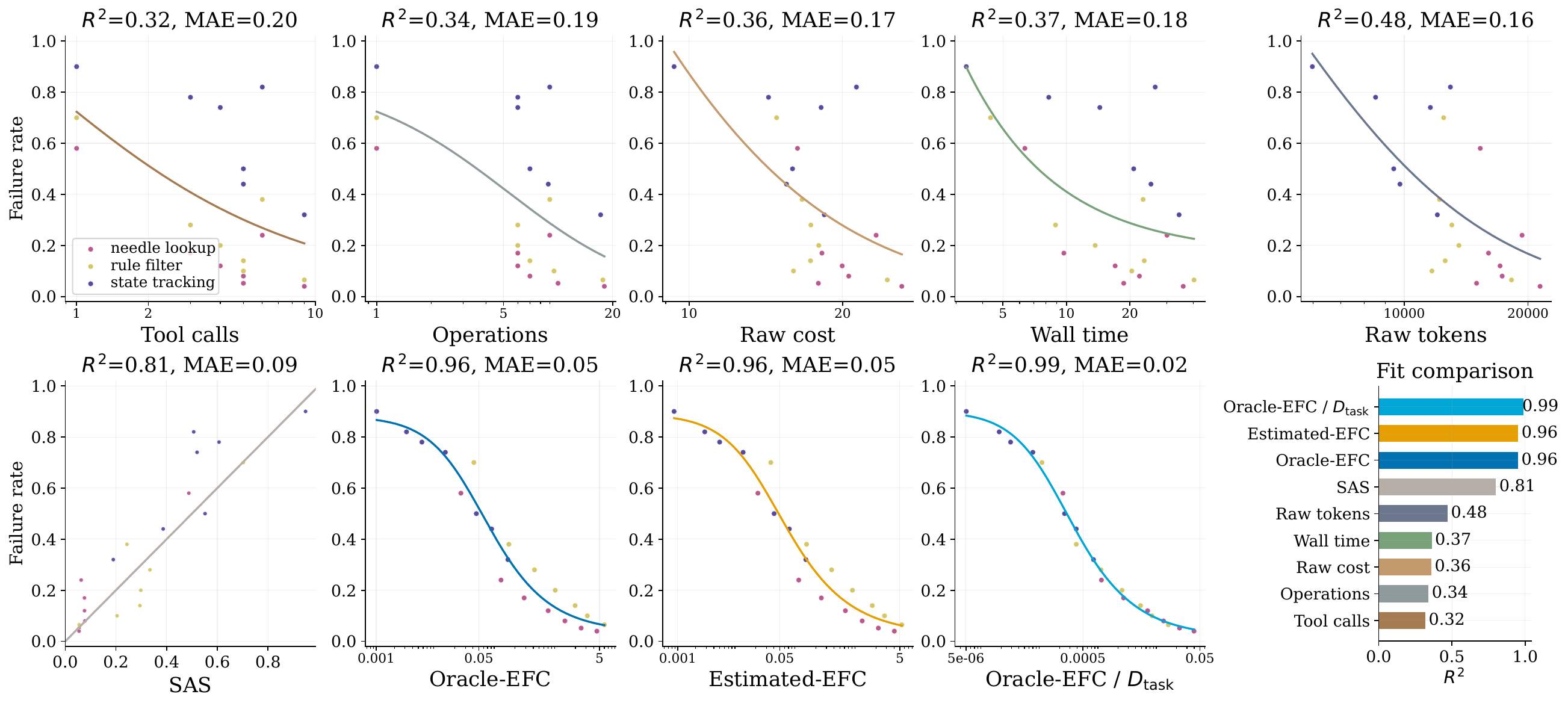}
    \caption{
    \textbf{Model-specific controlled scaling result for \texttt{gpt-5.4-nano}.}
    The analysis follows \S\ref{sec:controlled_scaling} but uses only
    trajectories from \texttt{gpt-5.4-nano}. Task-normalized Oracle-EFC gives
    the strongest fit, while Estimated-EFC closely matches Oracle-EFC.
    }
    \label{fig:controlled_scaling_gpt}
\end{figure*}

\begin{figure*}[t]
    \centering
    \includegraphics[width=.95\textwidth]{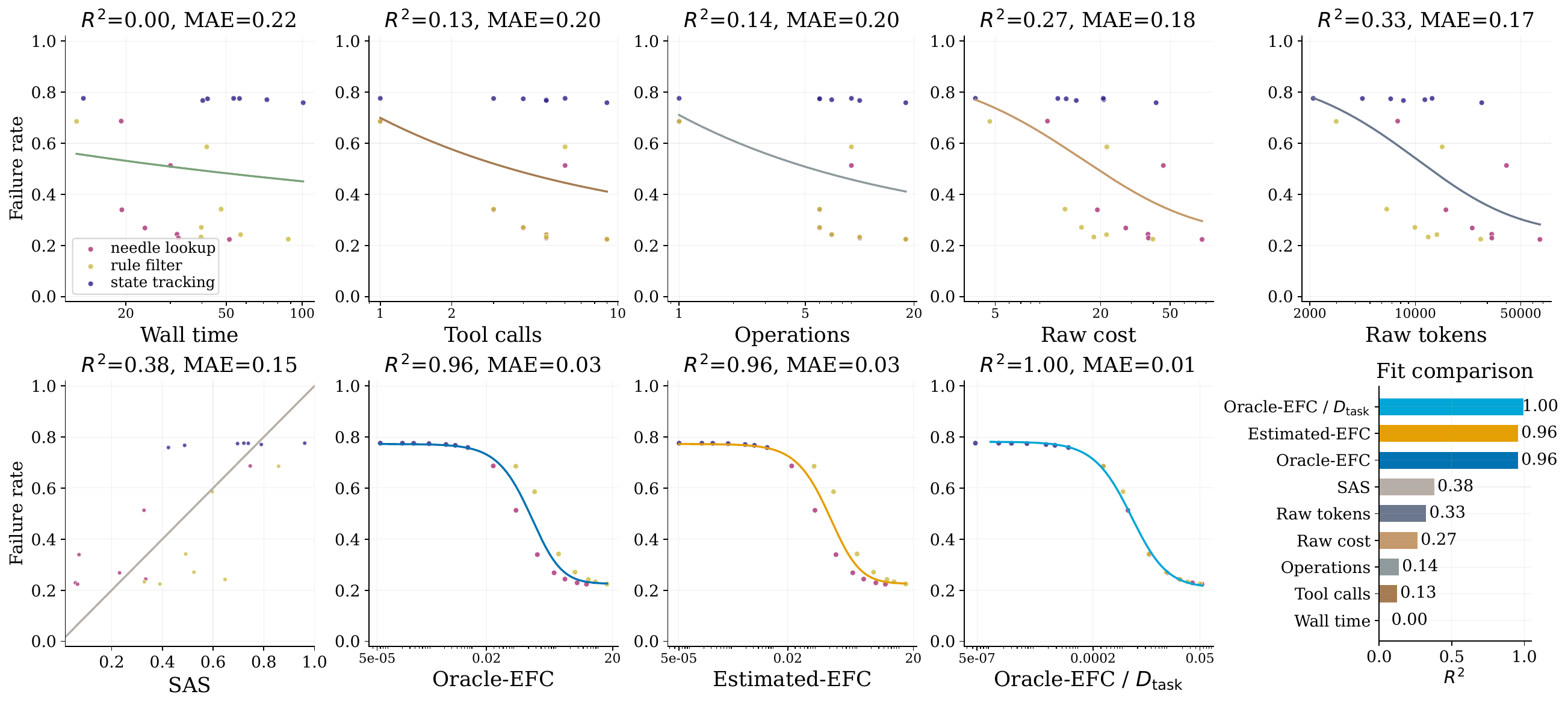}
    \caption{
    \textbf{Model-specific controlled scaling result for \texttt{DeepSeek-V4-Flash}.}
    The analysis follows \S\ref{sec:controlled_scaling} but uses only
    trajectories from \texttt{DeepSeek-V4-Flash}. EFC-based coordinates
    substantially outperform raw expenditure and SAS, and task normalization
    gives the best collapse.
    }
    \label{fig:controlled_scaling_dpsk}
\end{figure*}

\begin{figure*}[t]
    \centering
    \includegraphics[width=.95\textwidth]{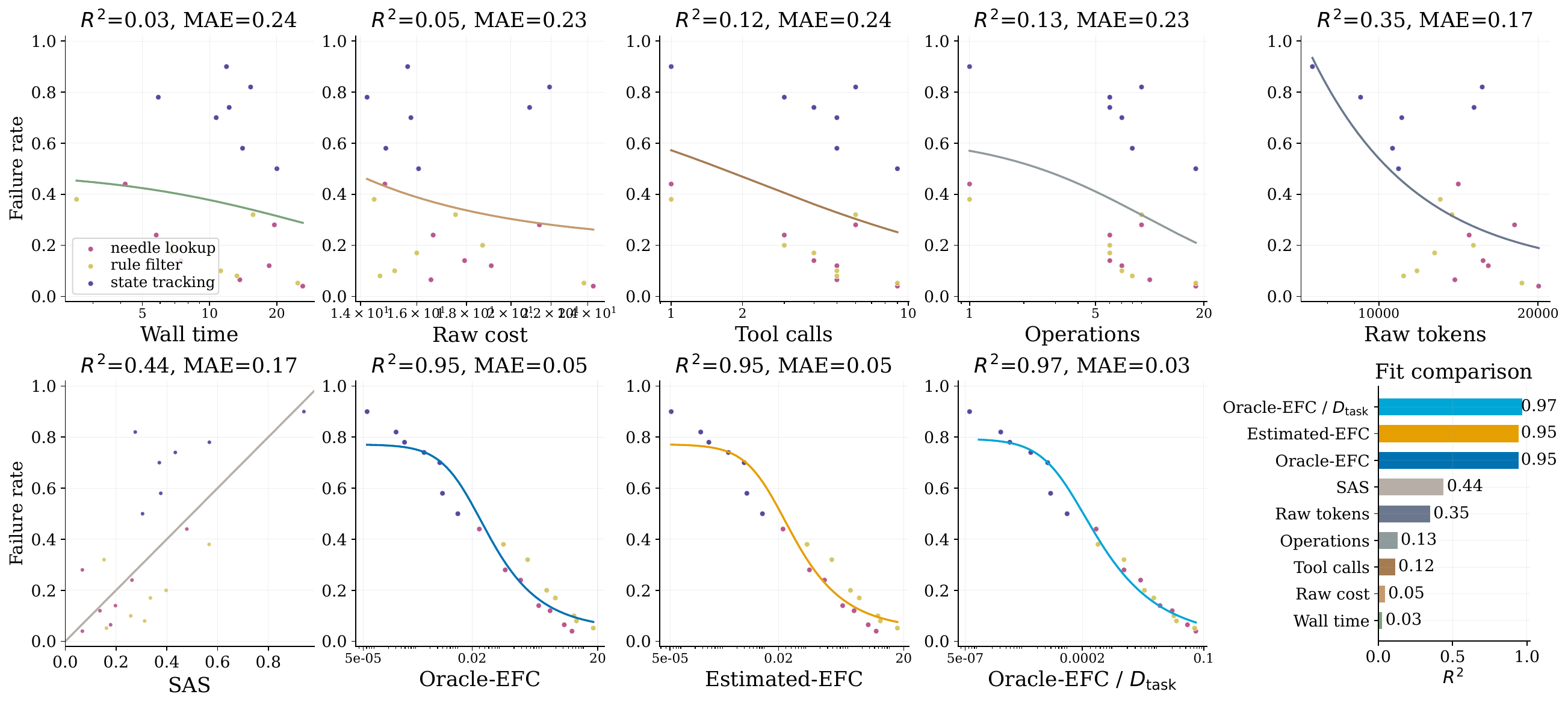}
    \caption{
    \textbf{Model-specific controlled scaling result for \texttt{Claude-Haiku-4.5}.}
    The analysis follows \S\ref{sec:controlled_scaling} but uses only
    trajectories from \texttt{Claude-Haiku-4.5}. Oracle-EFC and Estimated-EFC
    remain closely aligned, and task-normalized Oracle-EFC gives the strongest
    fit among all coordinates.
    }
    \label{fig:controlled_scaling_claude}
\end{figure*}

Figures~\ref{fig:controlled_scaling_gpt}--\ref{fig:controlled_scaling_claude}
repeat the controlled scaling analysis in \S\ref{sec:controlled_scaling}
separately for each base model.
The same ordering holds across models.
Raw expenditure measures provide only partial fits, with the best raw baseline
reaching $R^2=0.48$ for \texttt{gpt-5.4-nano}, $0.33$ for
\texttt{DeepSeek-V4-Flash}, and $0.35$ for \texttt{Claude-Haiku-4.5}.
SAS is stronger than most raw baselines, especially for \texttt{gpt-5.4-nano},
where it reaches $R^2=0.81$, but it remains below the EFC coordinates.
Oracle-EFC and Estimated-EFC closely match each other for all three models,
with $R^2$ between $0.95$ and $0.96$.
Normalizing Oracle-EFC by task demand gives the strongest collapse in every
case, reaching $R^2=0.99$ for \texttt{gpt-5.4-nano}, $1.00$ for
\texttt{DeepSeek-V4-Flash}, and $0.97$ for \texttt{Claude-Haiku-4.5}.
These results show that the aggregate trend in \S\ref{sec:controlled_scaling}
is not driven by a single backbone, and that task-normalized EFC remains the
most predictive coordinate across base-model capability levels.

\subsubsection{Model-Specific Trace-Time Estimation Results}
\label{app:model_specific_trace_estimation}

\begin{figure}[t]
    \centering
    \includegraphics[width=.95\linewidth]{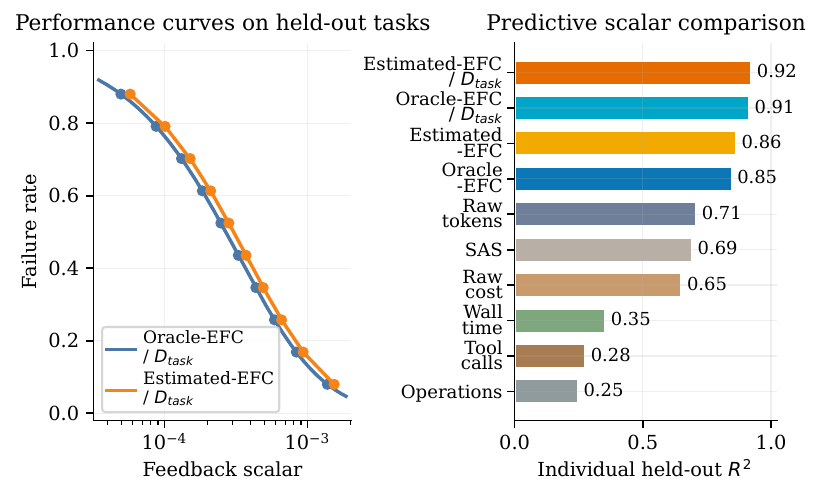}
    \caption{
    \textbf{Model-specific trace-time estimation result for \texttt{gpt-5.4-nano}.}
    The analysis follows \S\ref{sec:trace_estimation} but uses only
    trajectories from \texttt{gpt-5.4-nano}. Estimated-EFC closely matches
    Oracle-EFC, and Estimated-EFC/$D_{\mathrm{task}}$ gives the strongest
    held-out fit.
    }
    \label{fig:trace_estimation_gpt}
\end{figure}

\begin{figure}[t]
    \centering
    \includegraphics[width=.95\linewidth]{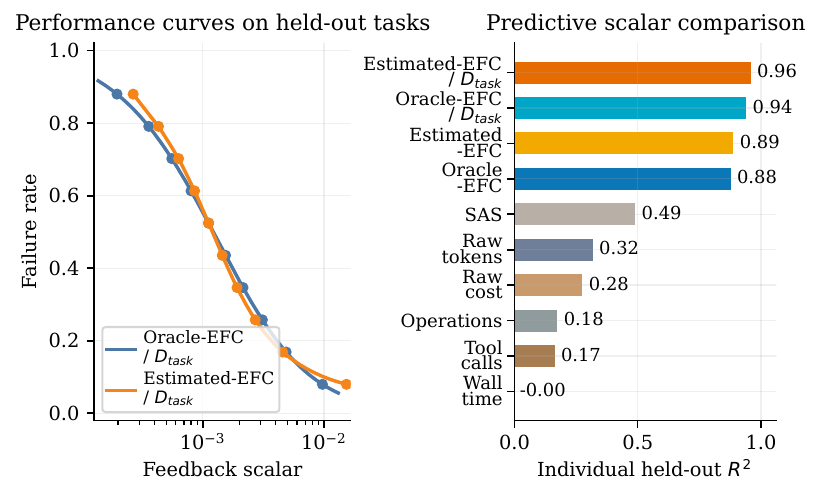}
    \caption{
    \textbf{Model-specific trace-time estimation result for \texttt{DeepSeek-V4-Flash}.}
    The analysis follows \S\ref{sec:trace_estimation} but uses only
    trajectories from \texttt{DeepSeek-V4-Flash}. The trace-time estimator
    slightly exceeds the oracle coordinate on this split after task
    normalization, while both EFC-based coordinates outperform raw expenditure
    and SAS.
    }
    \label{fig:trace_estimation_dpsk}
\end{figure}

\begin{figure}[t]
    \centering
    \includegraphics[width=.95\linewidth]{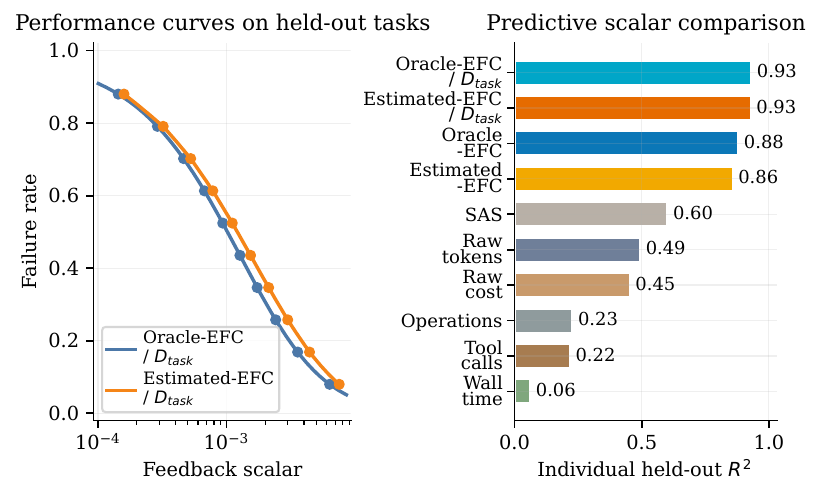}
    \caption{
    \textbf{Model-specific trace-time estimation result for \texttt{Claude-Haiku-4.5}.}
    The analysis follows \S\ref{sec:trace_estimation} but uses only
    trajectories from \texttt{Claude-Haiku-4.5}. Estimated-EFC and Oracle-EFC
    remain closely aligned, and their task-normalized variants give the best
    held-out fits.
    }
    \label{fig:trace_estimation_claude}
\end{figure}

Figures~\ref{fig:trace_estimation_gpt}--\ref{fig:trace_estimation_claude}
repeat the trace-time estimation analysis in \S\ref{sec:trace_estimation}
separately for each base model.
The same pattern holds across all three backbones.
Estimated-EFC closely tracks Oracle-EFC before task normalization, with
$R^2=0.86$ versus $0.85$ for \texttt{gpt-5.4-nano}, $0.89$ versus $0.88$ for
\texttt{DeepSeek-V4-Flash}, and $0.86$ versus $0.88$ for
\texttt{Claude-Haiku-4.5}.
Task normalization improves both coordinates in each case.
Estimated-EFC/$D_{\mathrm{task}}$ reaches $R^2=0.92$ for
\texttt{gpt-5.4-nano}, $0.96$ for \texttt{DeepSeek-V4-Flash}, and $0.93$ for
\texttt{Claude-Haiku-4.5}.
These values are consistently higher than raw-compute baselines and SAS, even
when raw tokens or SAS are relatively strong for a particular model.
The per-model results therefore support the conclusion in
\S\ref{sec:trace_estimation}: trace-time signals recover most of the oracle
feedback coordinate, and task demand removes residual scale mismatch across
held-out task families.

\subsubsection{Model-Specific Task-Demand Curve Collapse Results}
\label{app:model_specific_curve_collapse}

\begin{figure}[t]
    \centering
    \includegraphics[width=.8\columnwidth]{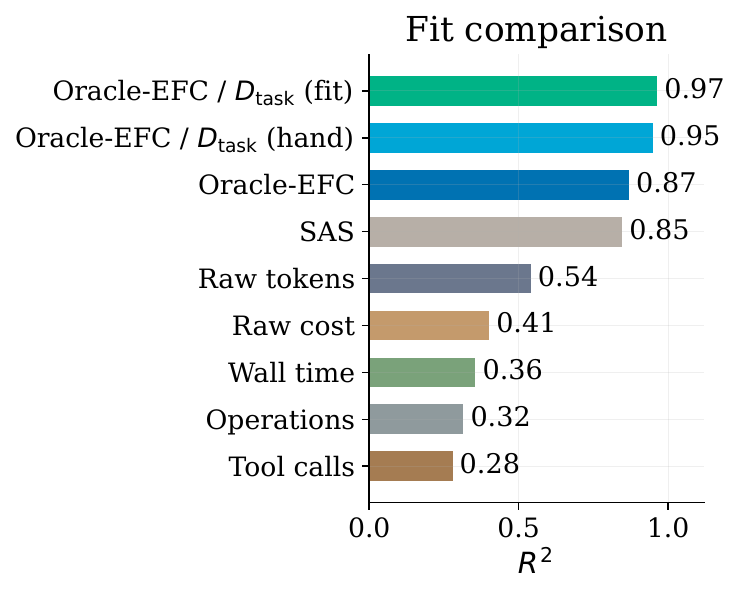}
    \caption{
    \textbf{Model-specific curve-collapse result for \texttt{gpt-5.4-nano}.}
    The analysis follows \S\ref{sec:difficulty_collapse} but uses only
    trajectories from \texttt{gpt-5.4-nano}. Task-normalized Oracle-EFC gives
    the strongest fit, with the fitted demand scale improving over the
    hand-designed demand scale.
    }
    \label{fig:curve_collapse_gpt}
\end{figure}

\begin{figure}[t]
    \centering
    \includegraphics[width=.8\columnwidth]{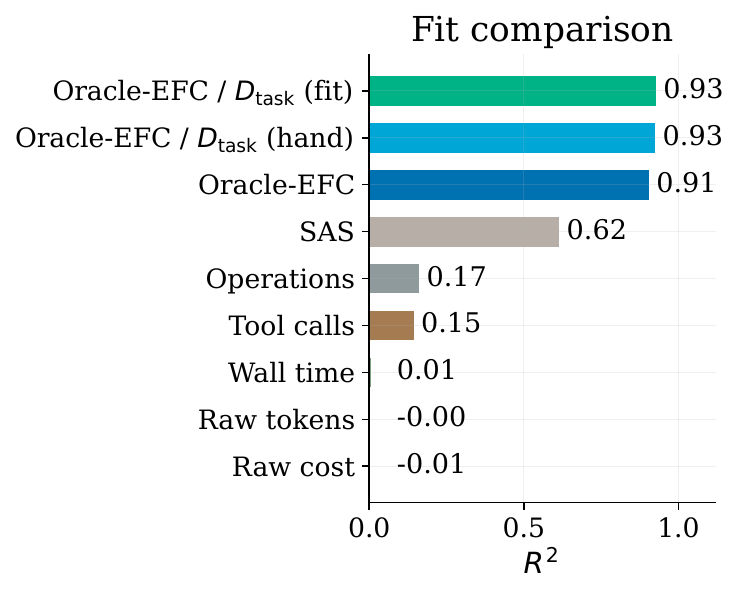}
    \caption{
    \textbf{Model-specific curve-collapse result for \texttt{DeepSeek-V4-Flash}.}
    The analysis follows \S\ref{sec:difficulty_collapse} but uses only
    trajectories from \texttt{DeepSeek-V4-Flash}. EFC-based coordinates
    outperform raw expenditure and SAS, and task-normalized Oracle-EFC gives
    the best fit.
    }
    \label{fig:curve_collapse_dpsk}
\end{figure}

\begin{figure}[t]
    \centering
    \includegraphics[width=.8\columnwidth]{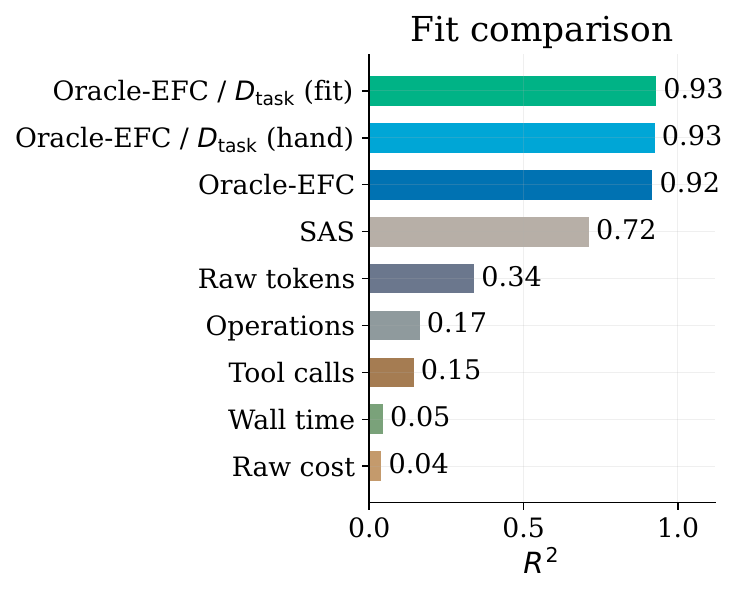}
    \caption{
    \textbf{Model-specific curve-collapse result for \texttt{Claude-Haiku-4.5}.}
    The analysis follows \S\ref{sec:difficulty_collapse} but uses only
    trajectories from \texttt{Claude-Haiku-4.5}. Oracle-EFC is already highly
    predictive, and task-demand normalization gives the strongest or
    tied-strongest fit.
    }
    \label{fig:curve_collapse_claude}
\end{figure}

Figures~\ref{fig:curve_collapse_gpt}--\ref{fig:curve_collapse_claude}
repeat the curve-collapse analysis in \S\ref{sec:difficulty_collapse}
separately for each base model.
The same conclusion holds across all three backbones.
Raw expenditure measures remain substantially weaker than feedback-based
coordinates.
For \texttt{gpt-5.4-nano}, the best raw baseline reaches $R^2=0.54$, while
SAS reaches $0.85$ and Oracle-EFC reaches $0.87$.
Task-demand normalization further improves the fit to $0.95$ with the
hand-designed demand scale and $0.97$ with the fitted demand scale.
For \texttt{DeepSeek-V4-Flash} and \texttt{Claude-Haiku-4.5}, raw baselines are
weaker, with best raw $R^2$ values of $0.17$ and $0.34$, respectively.
Oracle-EFC already gives strong fits on these models, reaching $0.91$ and
$0.92$, and both demand-normalized variants reach $0.93$.
These results show that the curve collapse reported in
\S\ref{sec:difficulty_collapse} is not an artifact of averaging across
models.
Task demand provides a stable scale for EFC across base-model capability
levels, and the fitted normalization mainly absorbs residual model-specific
scale mismatch.

\subsubsection{Model-Specific Task-Demand Calibration Transfer}
\label{app:model_specific_difficulty_transfer}

\begin{figure}[t]
    \centering
    \includegraphics[width=.95\columnwidth]{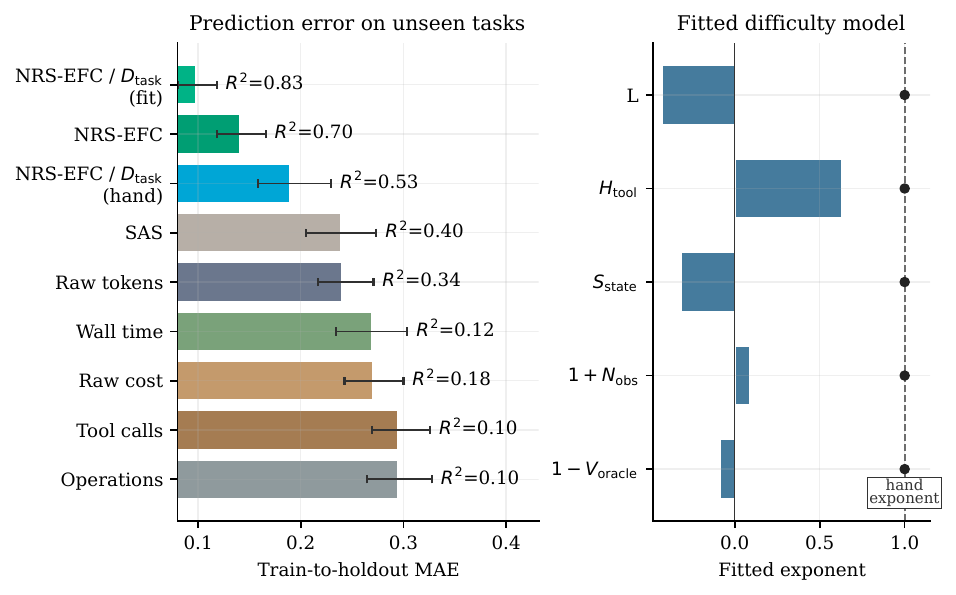}
    \caption{
    \textbf{Model-specific task-demand calibration transfer for
    \texttt{gpt-5.4-nano}.}
    The analysis follows \S\ref{sec:difficulty_transfer} but uses only
    trajectories from \texttt{gpt-5.4-nano}. The left panel compares
    train-to-holdout prediction error, and the right panel shows the fitted
    task-demand exponents.
    }
    \label{fig:difficulty_transfer_gpt}
\end{figure}

\begin{figure}[t]
    \centering
    \includegraphics[width=.95\columnwidth]{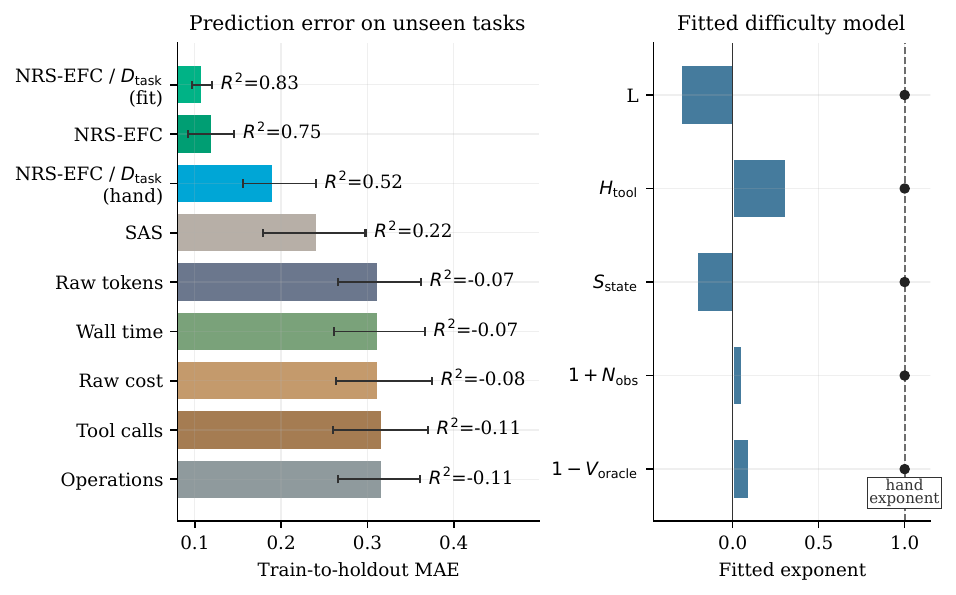}
    \caption{
    \textbf{Model-specific task-demand calibration transfer for
    \texttt{DeepSeek-V4-Flash}.}
    The analysis follows \S\ref{sec:difficulty_transfer} but uses only
    trajectories from \texttt{DeepSeek-V4-Flash}. The left panel compares
    train-to-holdout prediction error, and the right panel shows the fitted
    task-demand exponents.
    }
    \label{fig:difficulty_transfer_dpsk}
\end{figure}

\begin{figure}[t]
    \centering
    \includegraphics[width=.95\columnwidth]{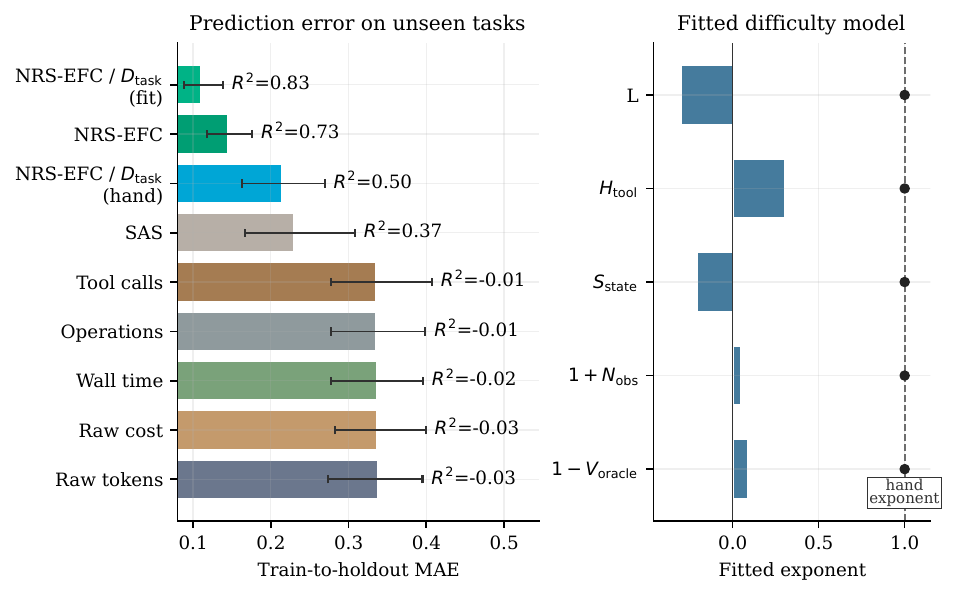}
    \caption{
    \textbf{Model-specific task-demand calibration transfer for
    \texttt{Claude-Haiku-4.5}.}
    The analysis follows \S\ref{sec:difficulty_transfer} but uses only
    trajectories from \texttt{Claude-Haiku-4.5}. The left panel compares
    train-to-holdout prediction error, and the right panel shows the fitted
    task-demand exponents.
    }
    \label{fig:difficulty_transfer_claude}
\end{figure}

Figures~\ref{fig:difficulty_transfer_gpt}--\ref{fig:difficulty_transfer_claude}
repeat the analysis in \S\ref{sec:difficulty_transfer} separately for each
base model. In the left panels, fitted task-demand normalization gives the
lowest train-to-holdout MAE and the highest held-out $R^2$ for all three
models. NRS-EFC/$D_{\mathrm{task}}$ with fitted demand reaches $R^2=0.83$ for
\texttt{gpt-5.4-nano}, \texttt{DeepSeek-V4-Flash}, and
\texttt{Claude-Haiku-4.5}. This improves over unnormalized NRS-EFC
($R^2=0.70$, $0.75$, and $0.73$) and over the hand-designed normalization
($R^2=0.53$, $0.52$, and $0.50$). SAS is weaker on all three models, and raw
expenditure baselines transfer poorly, with negative or near-zero $R^2$ values
for \texttt{DeepSeek-V4-Flash} and \texttt{Claude-Haiku-4.5}. In the right
panels, the fitted difficulty model consistently differs from the hand-designed
unit-exponent model. The fitted scale assigns a positive weight to
$H_{\mathrm{tool}}$, downweights or reverses the contribution of $L$ and
$S_{\mathrm{state}}$, and gives smaller weights to the observation-count and
oracle-validity terms. These per-model results support the conclusion in
\S\ref{sec:difficulty_transfer}: task-demand calibration improves transfer to
mixed held-out tasks, and the improvement is not driven by averaging across
base models.

\subsubsection{Model-Specific Prediction of Unseen Configurations}
\label{app:model_specific_heldout_prediction}

\begin{figure*}[t]
    \centering
    \includegraphics[width=.95\textwidth]{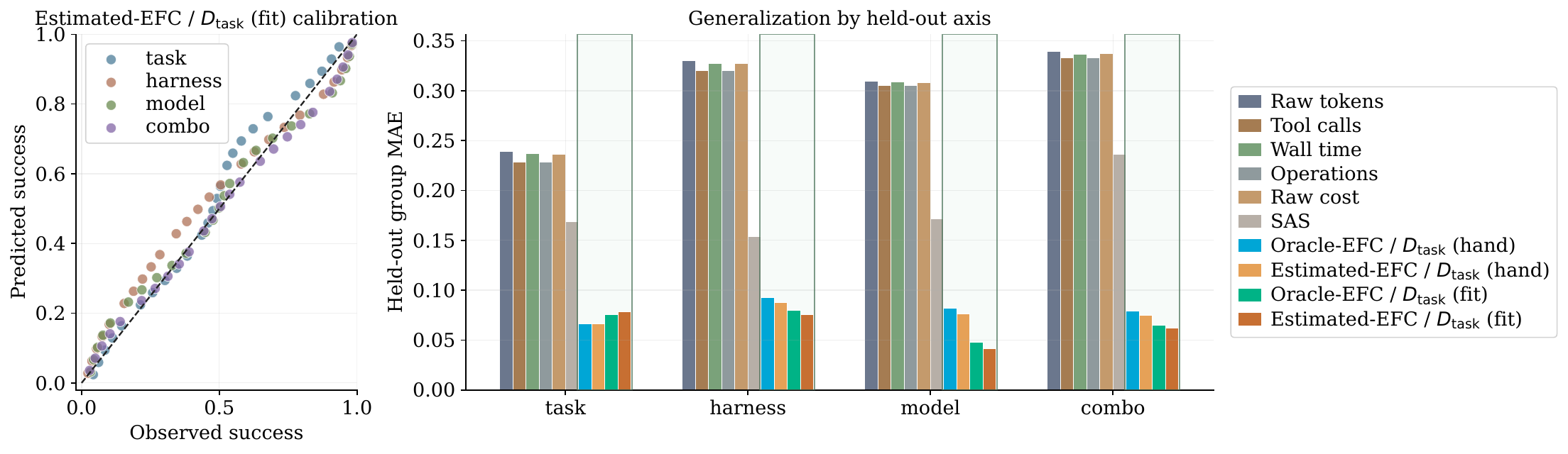}
    \caption{
    \textbf{Model-specific prediction of unseen configurations for
    \texttt{gpt-5.4-nano}.}
    The analysis follows \S\ref{sec:heldout_prediction} but uses only
    trajectories from \texttt{gpt-5.4-nano}. The left panel shows calibration
    between observed and predicted success under fitted
    Estimated-EFC/$D_{\mathrm{task}}$, and the right panel compares MAE across
    held-out task, harness, model, and combined splits.
    }
    \label{fig:heldout_prediction_gpt}
\end{figure*}

\begin{figure*}[t]
    \centering
    \includegraphics[width=.95\textwidth]{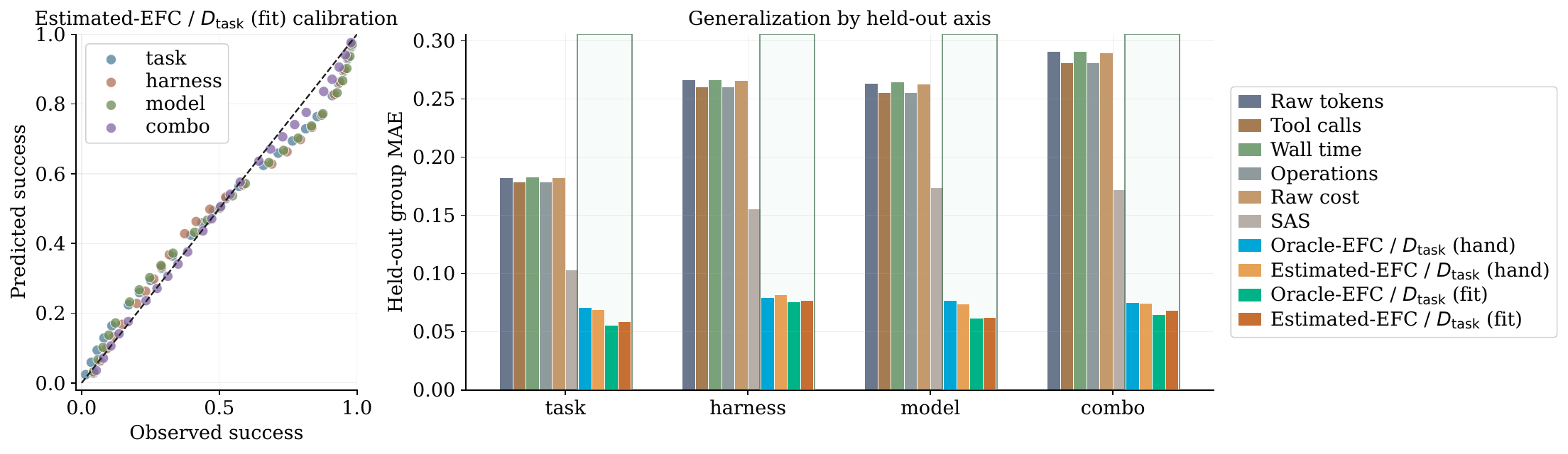}
    \caption{
    \textbf{Model-specific prediction of unseen configurations for
    \texttt{DeepSeek-V4-Flash}.}
    The analysis follows \S\ref{sec:heldout_prediction} but uses only
    trajectories from \texttt{DeepSeek-V4-Flash}. The left panel shows
    calibration between observed and predicted success under fitted
    Estimated-EFC/$D_{\mathrm{task}}$, and the right panel compares MAE across
    held-out task, harness, model, and combined splits.
    }
    \label{fig:heldout_prediction_dpsk}
\end{figure*}

\begin{figure*}[t]
    \centering
    \includegraphics[width=.95\textwidth]{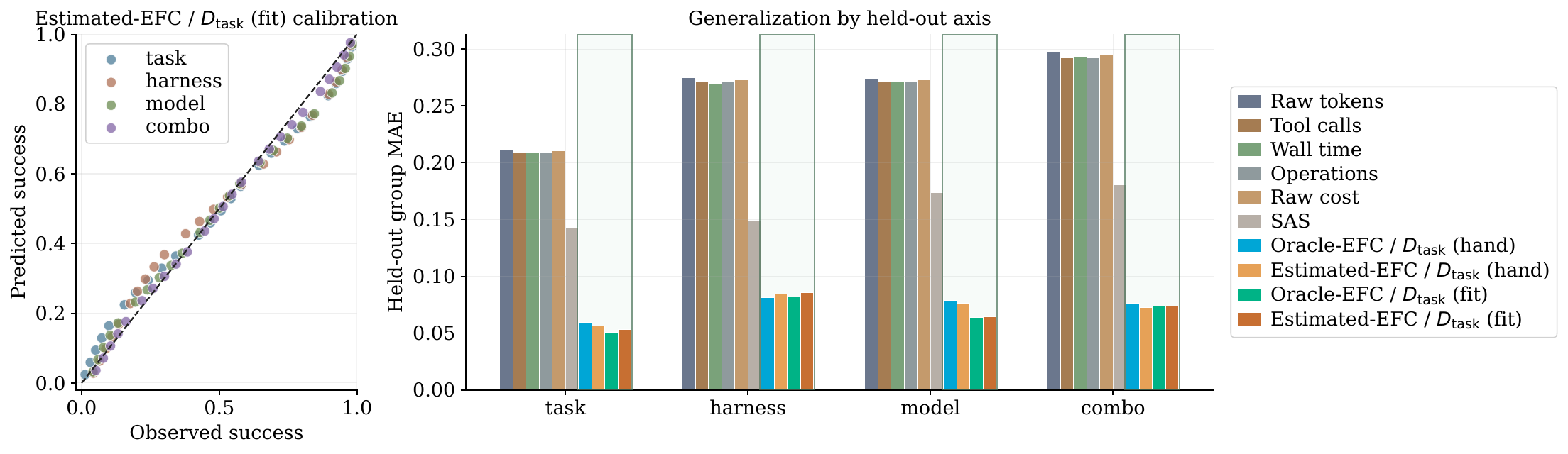}
    \caption{
    \textbf{Model-specific prediction of unseen configurations for
    \texttt{Claude-Haiku-4.5}.}
    The analysis follows \S\ref{sec:heldout_prediction} but uses only
    trajectories from \texttt{Claude-Haiku-4.5}. The left panel shows
    calibration between observed and predicted success under fitted
    Estimated-EFC/$D_{\mathrm{task}}$, and the right panel compares MAE across
    held-out task, harness, model, and combined splits.
    }
    \label{fig:heldout_prediction_claude}
\end{figure*}

Figures~\ref{fig:heldout_prediction_gpt}--\ref{fig:heldout_prediction_claude}
repeat the analysis in \S\ref{sec:heldout_prediction} separately for each base
model. In the left panels, fitted Estimated-EFC/$D_{\mathrm{task}}$ remains
well calibrated across task, harness, model, and combined held-out splits. The
points lie close to the diagonal for all three backbones, showing that the
learned EFC coordinate predicts absolute success rates rather than only ranking
configurations. In the right panels, raw expenditure measures have high MAE,
especially on held-out harness, model, and combined splits. SAS reduces this
error but remains substantially worse than task-normalized EFC. The fitted
Estimated-EFC/$D_{\mathrm{task}}$ coordinate gives low error across all axes,
with MAE roughly in the range of $0.04$--$0.08$ for most splits. The largest
gains appear on the model and combined axes, where raw-compute baselines often
exceed $0.25$ MAE. These results support the conclusion in
\S\ref{sec:heldout_prediction}: task-demand-normalized EFC generalizes to
unseen configurations within each base model, and the aggregate result is not
driven by cross-model averaging.

\subsubsection{Model-Specific Real-Mix Harness Efficiency}
\label{app:model_specific_real_mix}

\begin{figure}[t]
    \centering
    \includegraphics[width=.95\columnwidth]{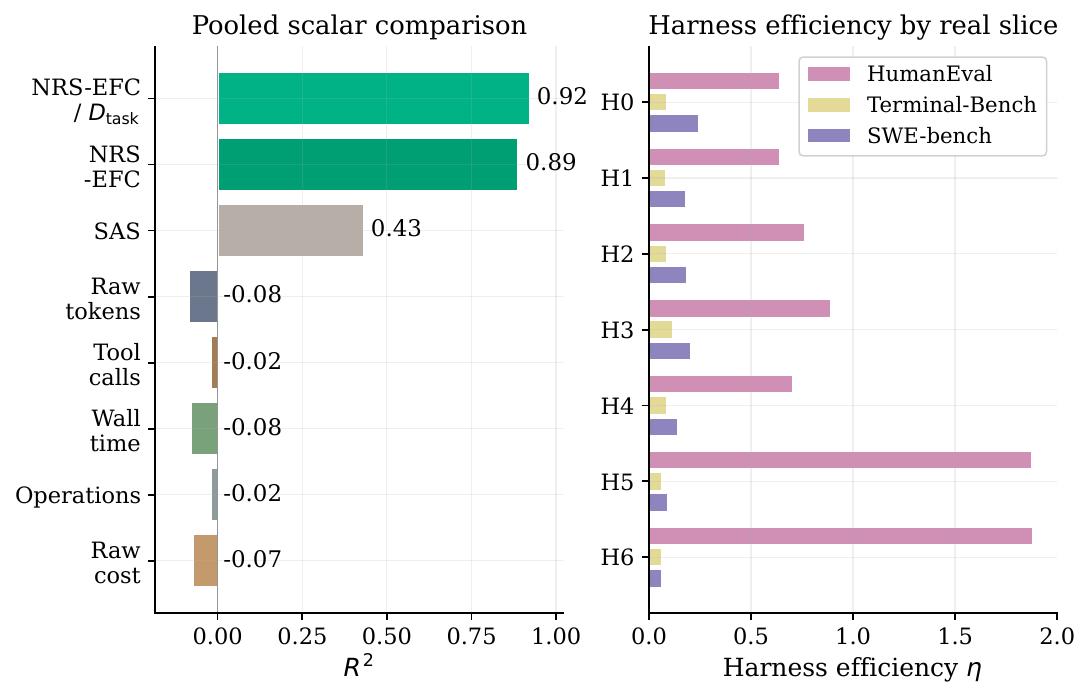}
    \caption{
    \textbf{Model-specific real-mix result for \texttt{gpt-5.4-nano}.}
    The analysis follows \S\ref{sec:real_mix} but uses only trajectories from
    \texttt{gpt-5.4-nano}. The left panel compares pooled scalar fits, and the
    right panel reports harness efficiency $\eta$ across real-task slices.
    }
    \label{fig:real_mix_gpt}
\end{figure}

\begin{figure}[t]
    \centering
    \includegraphics[width=.95\columnwidth]{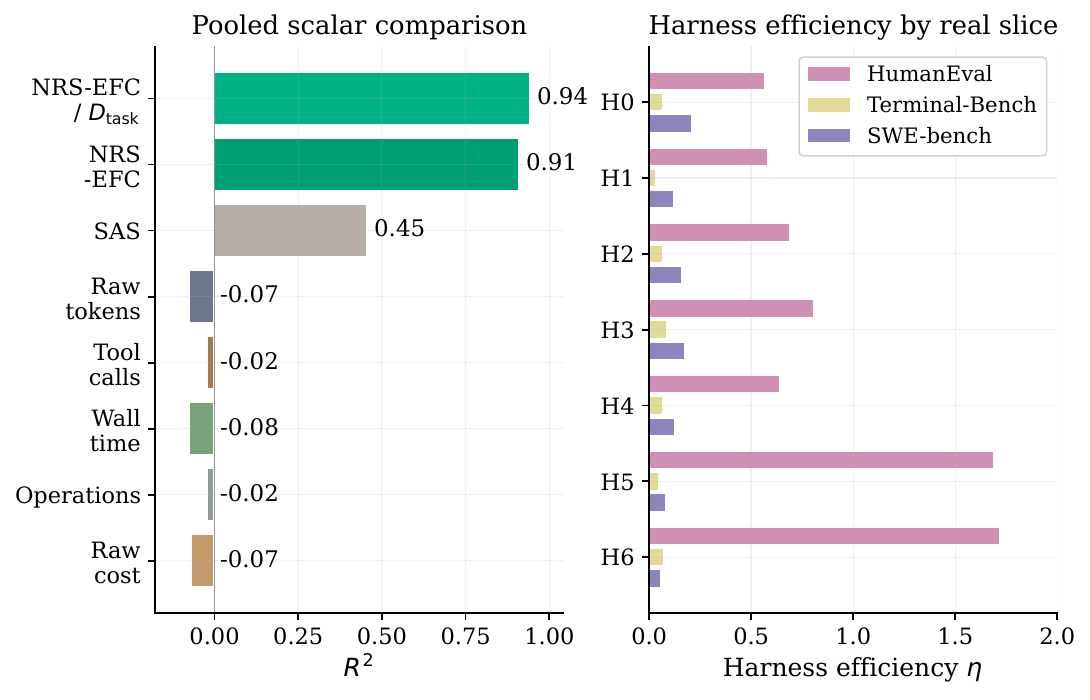}
    \caption{
    \textbf{Model-specific real-mix result for \texttt{DeepSeek-V4-Flash}.}
    The analysis follows \S\ref{sec:real_mix} but uses only trajectories from
    \texttt{DeepSeek-V4-Flash}. The left panel compares pooled scalar fits, and
    the right panel reports harness efficiency $\eta$ across real-task slices.
    }
    \label{fig:real_mix_dpsk}
\end{figure}

\begin{figure}[t]
    \centering
    \includegraphics[width=.95\columnwidth]{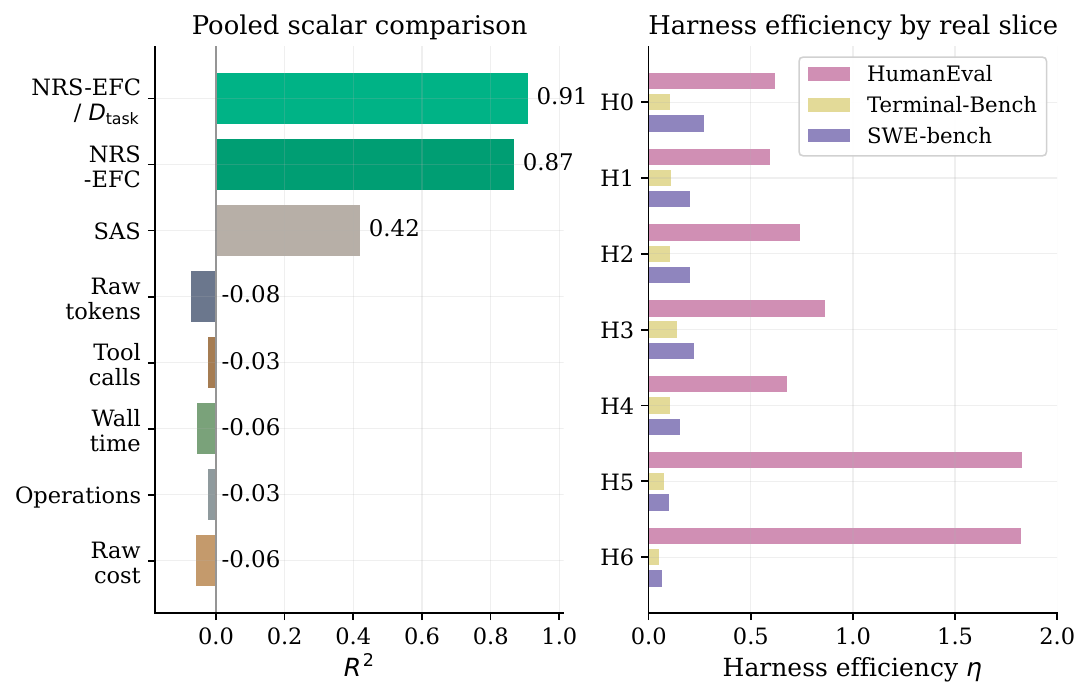}
    \caption{
    \textbf{Model-specific real-mix result for \texttt{Claude-Haiku-4.5}.}
    The analysis follows \S\ref{sec:real_mix} but uses only trajectories from
    \texttt{Claude-Haiku-4.5}. The left panel compares pooled scalar fits, and
    the right panel reports harness efficiency $\eta$ across real-task slices.
    }
    \label{fig:real_mix_claude}
\end{figure}

Figures~\ref{fig:real_mix_gpt}--\ref{fig:real_mix_claude} repeat the analysis
in \S\ref{sec:real_mix} separately for each base model. In the left panels,
NRS-EFC and NRS-EFC/$D_{\mathrm{task}}$ are consistently the strongest
predictors on pooled real traces. NRS-EFC/$D_{\mathrm{task}}$ reaches
$R^2=0.92$ for \texttt{gpt-5.4-nano}, $0.94$ for
\texttt{DeepSeek-V4-Flash}, and $0.91$ for \texttt{Claude-Haiku-4.5}.
Unnormalized NRS-EFC is slightly weaker but remains strong, with $R^2=0.89$,
$0.91$, and $0.87$, respectively. SAS gives moderate fits around
$R^2=0.42$--$0.45$, while raw expenditure measures have near-zero or negative
$R^2$ values. In the right panels, the harness-efficiency profiles are also
stable across models. HumanEval exhibits the largest $\eta$ values, with H5
and H6 producing the strongest efficiency gains. Terminal-Bench and SWE-bench
show smaller absolute efficiencies, indicating that additional raw expenditure
is converted into useful non-redundant feedback less efficiently on these
harder real-task slices. These results support the conclusion in
\S\ref{sec:real_mix}: NRS-EFC identifies slice-specific harness efficiency, and
the real-trace trends are not driven by averaging across base models.

\subsubsection{Model-Specific Prospective Holdout Results}
\label{app:model_specific_prospective_holdout}

\begin{figure}[t]
    \centering
    \includegraphics[width=.85\linewidth]{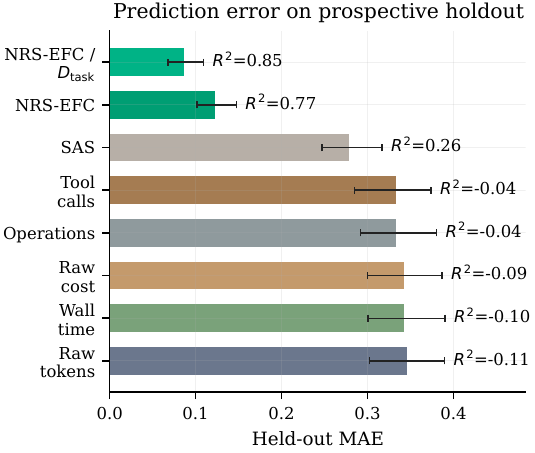}
    \caption{
    \textbf{Model-specific prospective holdout result for
    \texttt{gpt-5.4-nano}.}
    The analysis follows \S\ref{sec:prospective_holdout} but uses only
    trajectories from \texttt{gpt-5.4-nano}. Bars report held-out MAE, and
    labels report the corresponding held-out $R^2$.
    }
    \label{fig:prospective_holdout_gpt}
\end{figure}

\begin{figure}[t]
    \centering
    \includegraphics[width=.85\linewidth]{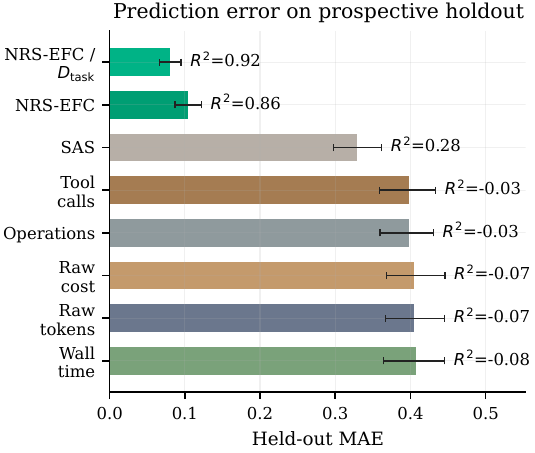}
    \caption{
    \textbf{Model-specific prospective holdout result for
    \texttt{DeepSeek-V4-Flash}.}
    The analysis follows \S\ref{sec:prospective_holdout} but uses only
    trajectories from \texttt{DeepSeek-V4-Flash}. Bars report held-out MAE, and
    labels report the corresponding held-out $R^2$.
    }
    \label{fig:prospective_holdout_dpsk}
\end{figure}

\begin{figure}[t]
    \centering
    \includegraphics[width=.85\linewidth]{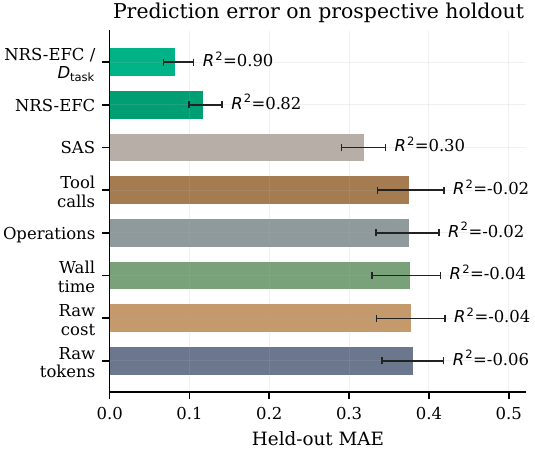}
    \caption{
    \textbf{Model-specific prospective holdout result for
    \texttt{Claude-Haiku-4.5}.}
    The analysis follows \S\ref{sec:prospective_holdout} but uses only
    trajectories from \texttt{Claude-Haiku-4.5}. Bars report held-out MAE, and
    labels report the corresponding held-out $R^2$.
    }
    \label{fig:prospective_holdout_claude}
\end{figure}

Figures~\ref{fig:prospective_holdout_gpt}--\ref{fig:prospective_holdout_claude}
repeat the prospective validation in \S\ref{sec:prospective_holdout}
separately for each base model. The same ordering appears across all three
backbones. NRS-EFC/$D_{\mathrm{task}}$ gives the lowest held-out MAE and the
highest held-out $R^2$, reaching $R^2=0.85$ for \texttt{gpt-5.4-nano},
$0.92$ for \texttt{DeepSeek-V4-Flash}, and $0.90$ for
\texttt{Claude-Haiku-4.5}. NRS-EFC is consistently weaker but
still predictive for all three models. SAS provides only a
moderate signal, with $R^2$ between $0.26$ and $0.30$. Raw expenditure measures
do not transfer to the prospective holdout, yielding negative
$R^2$ values for tool calls, operations, wall time, raw cost, and raw tokens.
These results support the conclusion in \S\ref{sec:prospective_holdout}:
prospective generalization depends on retained, non-redundant, and
task-normalized feedback rather than on raw test-time expenditure, and this
pattern is stable within each base model.

\section{\ours: Full Method Description}
\label{app:efc_adapter}
We use EFC not only as an analysis coordinate, but also as a lightweight
companion signal for existing agent harnesses. We propose \ours, an
EFC-aware auxiliary layer that can be attached to a test-time scaling,
reflection, verifier-guided, memory-based, self-evolving, or multi-agent
harness. The adapter does not replace the base method. It keeps the same base
model, task interface, tool set, action space, and raw-budget limit, and only
adds feedback accounting, feedback selection, memory/evolution gating, and
stopping control. Its goal is to improve the conversion from raw expenditure to
effective feedback, namely to increase
$\eta=\mathrm{EFC}/C_{\mathrm{raw}}$, especially for demand-normalized
NRS-EFC.

\paragraph{Inputs and invariants.}
For a task instance $x$, a base harness $h$, and a raw budget $b$, the base
method produces a trajectory $\tau=\{(s_t,a_t,o_t,u_t)\}_{t=1}^{T}$ as defined
in \S\ref{sec:closed_loop}. \ours receives the task-visible input,
the base-method trace, the raw-budget accounting variables, and frozen EFC and
task-demand estimators. It does not use hidden solutions or final oracle success
when scoring feedback events. The final success label is reserved for evaluation.
Before execution, the adapter estimates a task-demand scale
$\widehat{D}_{\mathrm{task}}$ from task-visible features such as expected
interaction length, tool ambiguity, state pressure, observation ambiguity, and
available verifier signals. The primary control coordinate is
\begin{equation}
\begin{aligned}
\widehat{X}^{\mathrm{nr}}(\tau)
=
\frac{
\sum_{t=1}^{T}
\widehat{\mathrm{EFC}}^{\mathrm{nr}}_t
}{
\widehat{D}_{\mathrm{task}}
},
\\
\widehat{\eta}^{\mathrm{nr}}(\tau)
=
\frac{
\sum_{t=1}^{T}
\widehat{\mathrm{EFC}}^{\mathrm{nr}}_t
}{
C_{\mathrm{raw}}(\tau)
}.
\label{eq:adapter_nrs_objective}
\end{aligned}
\end{equation}
Thus, the adapter favors trajectories that produce stable, non-redundant
feedback at low raw cost, rather than trajectories that merely spend more
tokens, tool calls, or attempts.

\paragraph{Feedback accounting.}
During execution, \ours observes the base harness in a trace-local manner.
Each tool call, verifier result, reflection, memory update, rollout summary,
candidate comparison, repair step, or coordination message is represented as a
feedback event. The adapter records whether the event introduced new
task-relevant information, whether the signal was grounded in reliable evidence,
whether it was non-redundant with earlier feedback, whether it changed the
subsequent plan or candidate solution, and how much raw cost it consumed. 
For online control, \ours scores each feedback event with a prefix-only NRS-EFC
proxy, emphasizing feedback that is currently valid, non-redundant relative to
the prefix, incorporated into the current plan or memory, and not merely a
repeated attempt.
This produces an EFC ledger for the run:
\begin{equation}
\mathcal{L}(\tau)
=
\left\{
(e_t,\widehat{\mathrm{EFC}}^{\mathrm{nr}}_t,
\Delta C_t,\widehat{X}^{\mathrm{nr}}_t)
\right\}_{t=1}^{T}.
\label{eq:efc_ledger}
\end{equation}

\paragraph{Feedback selection.}
When the base method exposes multiple possible feedback-producing actions, such
as candidate rollouts, tests, tool queries, verifier calls, reflection prompts,
or prior trajectories to reuse, \ours ranks them by expected marginal
feedback return:
\begin{equation}
\mathrm{ROI}(a_t)
=
\frac{
\mathbb{E}\!\left[
\Delta \widehat{\mathrm{EFC}}^{\mathrm{nr}}(a_t)
/\widehat{D}_{\mathrm{task}}
\mid \tau_{<t}, x
\right]
}{
\mathbb{E}\!\left[
\Delta C_{\mathrm{raw}}(a_t)
\mid \tau_{<t}, x
\right]
}.
\label{eq:adapter_roi}
\end{equation}
The expectations in Eq.~\ref{eq:adapter_roi} are estimated with the prefix-only
estimator trained on calibration traces. No feature that requires future events,
later references, later memory retention, hidden solutions, or final success
labels is used when the adapter ranks actions during execution.
The adapter therefore prefers actions expected to reduce a real uncertainty,
produce targeted verifier evidence, localize an error, or change the next
decision. It downweights repeated reflection, unchanged test reruns, redundant
searches, and similar retries that increase raw cost without adding stable
feedback. If the base method does not expose alternative actions at a step, the
adapter only audits the event and uses the score for later memory, evolution, or
stopping decisions.

\paragraph{Memory and self-evolution gating.}
For reflection and memory-based harnesses, \ours treats memory as a scarce durable resource rather than a transcript of all observations. 
High-NRS-EFC feedback is promoted to durable memory when it is verified,
non-redundant relative to the current prefix, and incorporated into the current
plan or candidate solution. Plausible but unverified critiques remain in
temporary scratch space.
Repeated, invalid, or decision-irrelevant observations are discarded or assigned lower retrieval weight. 
For self-evolving agents, the same event-level credit assignment gates updates to memory, retrievers, prompts, policies, verifier rubrics, or skill libraries. 
In this view, self-evolution is not driven only by final task success or coarse reward; it is driven by which feedback events reliably improved future raw-to-EFC conversion. 
The evolution objective is therefore to increase future $\widehat{\eta}^{\mathrm{nr}}$ and $\widehat{X}^{\mathrm{nr}}$, not to store more experience.

\paragraph{Stopping and rollback.}
\ours also prevents low-value loops. Let $W$ be a recent window of
trajectory steps. If raw cost continues to grow while marginal normalized
NRS-EFC remains flat,
\begin{equation}
\frac{
\sum_{t\in W}\Delta \widehat{\mathrm{EFC}}^{\mathrm{nr}}_t
/\widehat{D}_{\mathrm{task}}
}{
\sum_{t\in W}\Delta C_t
}
< \epsilon,
\label{eq:adapter_stopping}
\end{equation}
the adapter stops, rolls back to a higher-scoring candidate, or switches to a
different base-method branch when such an option is already available. This
criterion distinguishes cases where the budget is not exhausted but useful
feedback has saturated from cases where additional cost is still producing
stable evidence.
The window size $W$ and threshold $\epsilon$ are fixed on the calibration split
and then kept unchanged for all reported adapter evaluations.

\paragraph{Compatibility with existing harnesses.}
\ours is designed as a companion layer rather than a competing harness. In
test-time scaling and trajectory-refinement methods, it can rank already
generated rollouts, select prior summaries for refinement, or choose the final
candidate using EFC-aware scores. In verifier- or PRM-guided methods, it filters
which verifier signals should trigger repair or memory updates. In reflection
and self-refine methods, it gates which critiques are retained and reused. In
self-evolving agents, it provides event-level credit assignment for deciding
what to evolve, when to evolve, and whether the evolution improved feedback
efficiency. In multi-agent settings, the same accounting can be applied to
critic messages, votes, cross-checks, and coordinator decisions.


\paragraph{Matched-budget evaluation.}
To ensure that gains do not come from hidden extra computation, we evaluate
\ours under matched raw-budget caps. The base method and its EFC-assisted
variant use the same model, decoding settings, task split, prompt template,
tool set, tool permissions, action space, and maximum raw-budget caps. The
adapter may only rerank, retain, discard, gate, stop, or roll back within the
feedback actions already available to the base harness. Because the adapter may
stop early, the comparison matches budget caps rather than realized expenditure, 
we therefore report realized raw cost separately. Under this protocol, an
improvement over the base method is attributed to better raw-to-feedback
conversion rather than to spending more allowed compute.

\end{document}